\documentclass[sn-mathphys-num]{sn-jnl}

\usepackage{graphicx}%
\usepackage{multirow}%
\usepackage{amsmath,amssymb,amsfonts}%
\usepackage{amsthm}%
\usepackage{mathrsfs}%
\usepackage[title]{appendix}%
\usepackage{xcolor}%
\usepackage{textcomp}%
\usepackage{manyfoot}%
\usepackage{booktabs}%
\usepackage{algorithm}%
\usepackage{subcaption}
\usepackage{listings}%
\usepackage{adjustbox}

\usepackage{times}
\usepackage{soul}
\usepackage{url}
\usepackage{graphicx}
\usepackage{algorithmic}
\usepackage[switch]{lineno}
\usepackage{arydshln}
\usepackage{rotating}

\newtheorem{definition}{Definition}%

\def\method{\texttt{ATNPA}}
\DeclareMathOperator*{\hadmard}{\odot}

\begin{document}


\title[Article Title]{Oversmoothing Alleviation in Graph Neural Networks: A Survey and Unified View}


\author*[1]{\fnm{Yufei} \sur{Jin}}\email{yjin2021@fau.edu}

\author[2]{\fnm{Xingquan} \sur{Zhu}}\email{xzhu3@fau.edu}
\equalcont{These authors contributed equally to this work.}


\affil[1]{\orgdiv{ Dept. of Electrical Engineering \& Computer Science}, \orgname{Florida Atlantic University}, \orgaddress{\city{Boca Raton}, \postcode{33431}, \state{FL}, \country{USA}}}

\affil[2]{\orgdiv{ Dept. of Electrical Engineering \& Computer Science}, \orgname{Florida Atlantic University}, \orgaddress{\city{Boca Raton}, \postcode{33431}, \state{FL}, \country{USA}}}



\abstract{Oversmoothing is a common challenge in learning graph neural networks (GNN), where, as layers increase, embedding features learned from GNNs quickly become similar or indistinguishable, making them incapable of differentiating network proximity. A GNN with shallow layer architectures can only learn short-term relation or localized structure information, limiting its power of learning long-term connection, evidenced by their inferior learning performance on heterophilous graphs. Tackling oversmoothing is crucial for harnessing deep-layer architectures for GNNs. To date, many methods have been proposed to alleviate oversmoothing. The vast difference behind their design principles, combined with graph complications, make it difficult to understand and even compare the difference between different approaches in tackling the oversmoothing. In this paper, we propose \method, a unified view with five key steps: Augmentation, Transformation, Normalization, Propagation, and Aggregation, to summarize GNN oversmoothing alleviation approaches. We first propose a taxonomy for GNN oversmoothing alleviation which includes three themes to tackle oversmoothing. After that, we separate all methods into six categories, followed by detailed reviews of representative methods, including their relation to \method, and discussion of their niche, strength, and weakness. The review not only draws an in-depth understanding of existing methods in the field but also shows a clear road map for future study. 

}

\keywords{Oversmoothing, graph neural networks, graph embedding, graph learning, review}



\maketitle

\section{Introduction}\label{sec1}

Graph Neural Networks (GNN)~\cite{Gori2005graph,bruna2014spectral,defferrard2016convolutional} have become prevalent in support learning from networked data, especially after the success of the Graph Convolution Network (GCN)~\cite{Kipf2017gcn}. The main goal of GNN is to learn feature representation~\cite{Zhang2017NetworkRL} for network entities, such as nodes or edges, in order to support downstream tasks, like node classification or link prediction. While GNN has achieved competitive performance in many benchmark graph datasets, it is known to only perform well with shallow layer architectures but cannot learn long-term node-node relation well. One consequence of such inability leads to its inferior performance on heterophilous graph~\cite{zhu2021graph}. 

\begin{figure}
    \centering
    \begin{minipage}{0.32\textwidth}
    \centering
    \subcaptionbox{Karate Club Topology\label{fig:topology}}{
    \includegraphics[width=0.9\textwidth]{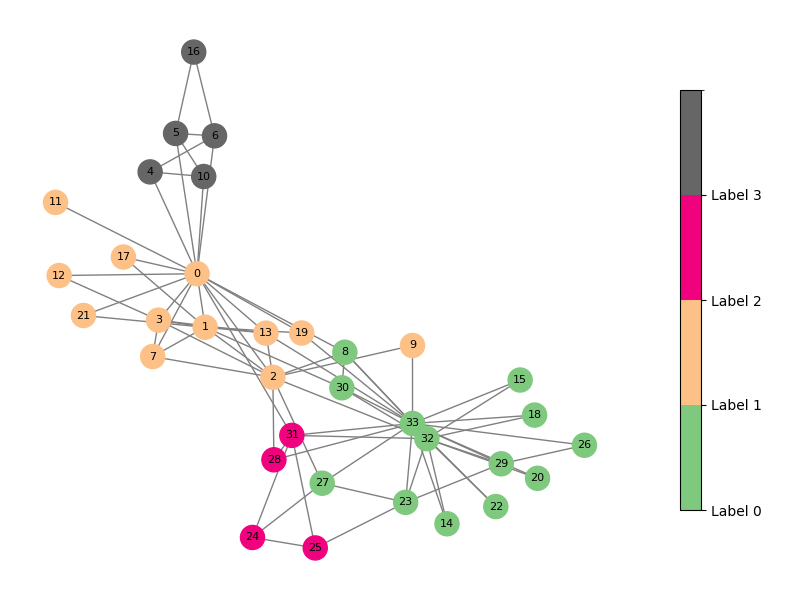}}
    \end{minipage} \hfill
    \begin{minipage}{0.32\textwidth}
    \centering
    \subcaptionbox{1-layer GCN Embedding\label{fig:k1}}{
    \includegraphics[width=0.9\textwidth]{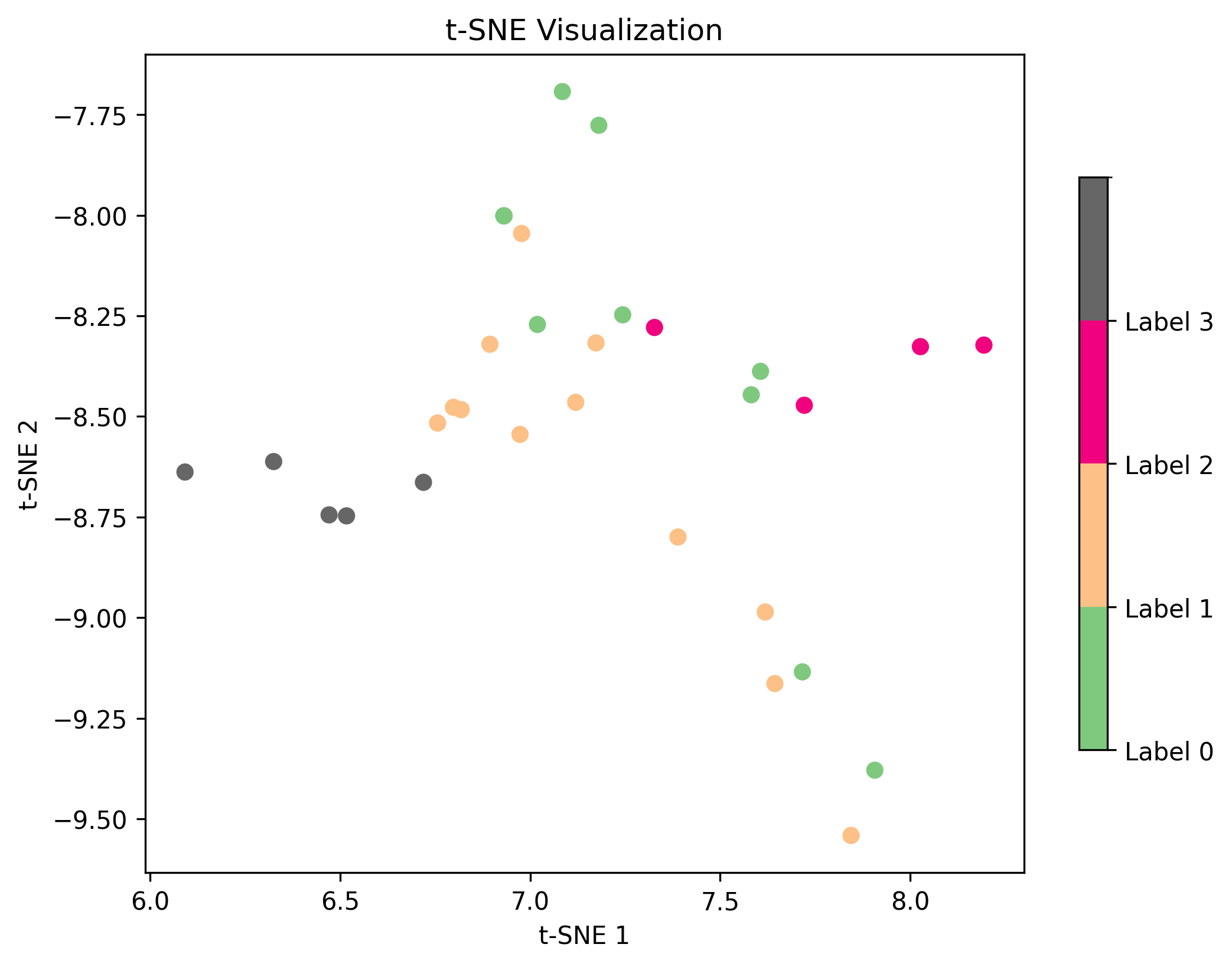}}
    \end{minipage} \hfill
    \begin{minipage}{0.32\textwidth}
    \centering
    \subcaptionbox{3-layers GCN Embedding\label{fig:k3}}{
    \includegraphics[width=0.9\textwidth]{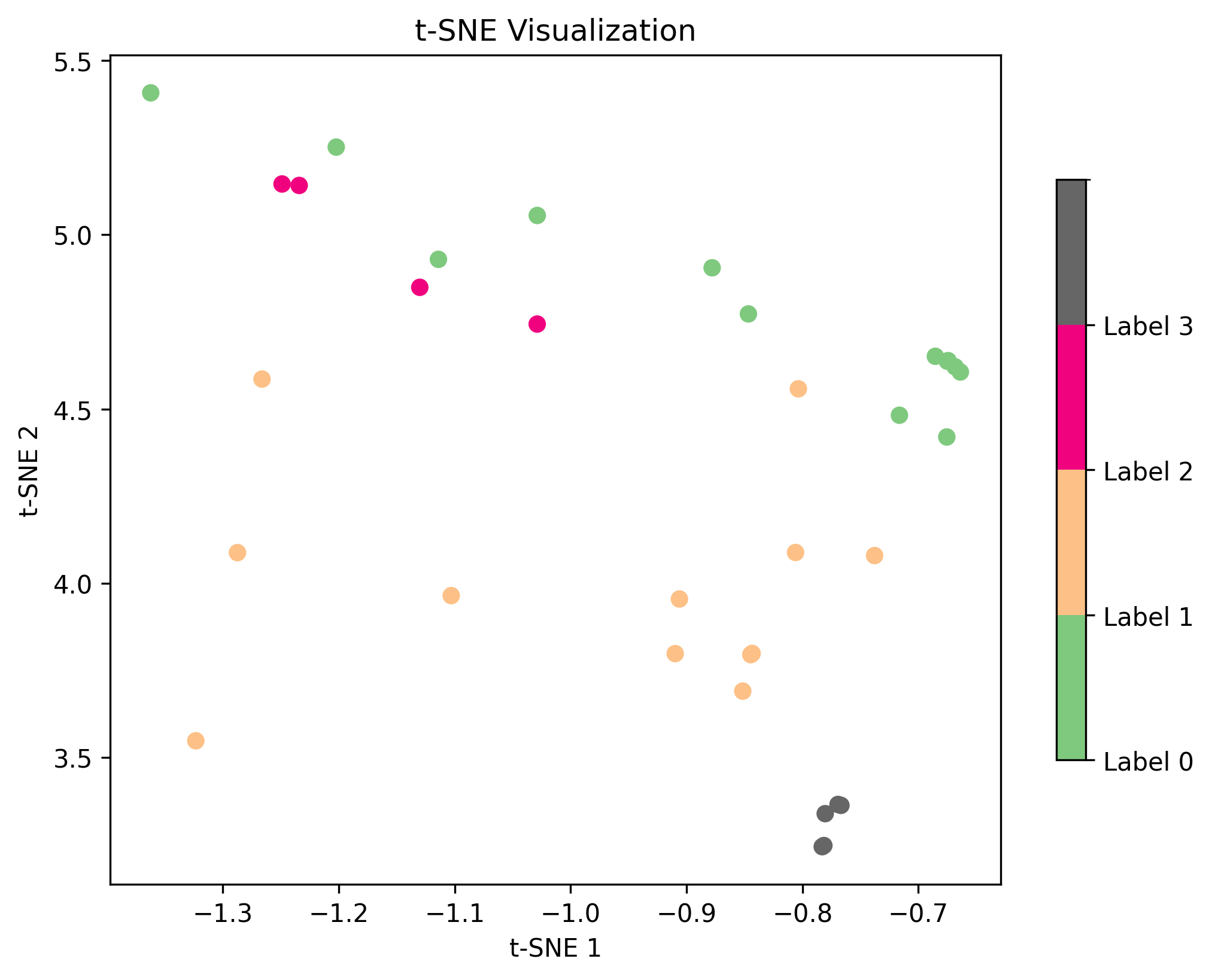}}
    \end{minipage} \hfill
    \begin{minipage}{0.32\textwidth}
    \centering
    \subcaptionbox{5-layers GCN Embedding\label{fig:k5}}{
    \includegraphics[width=0.9\textwidth]{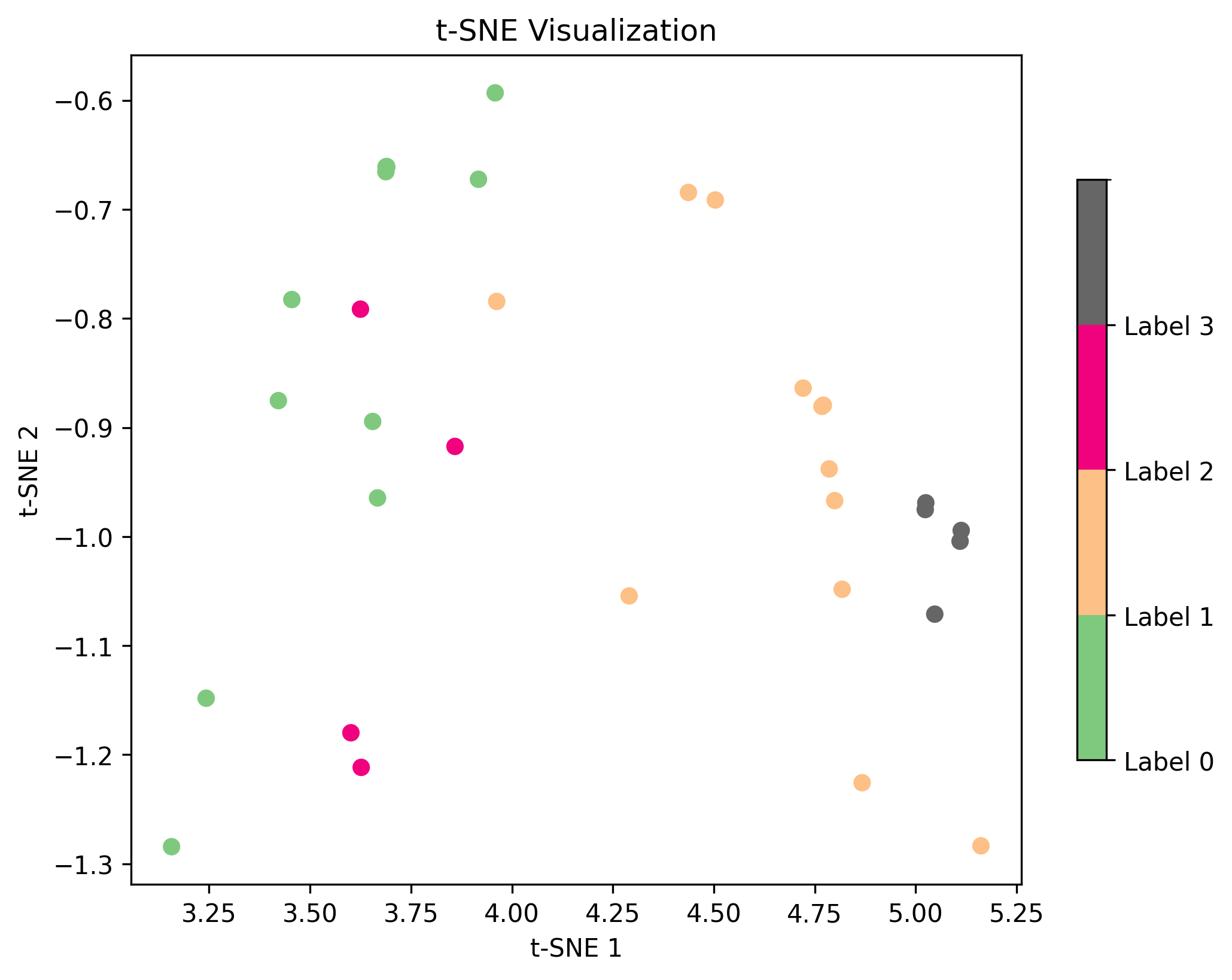}}
    \end{minipage} \hfill
    \begin{minipage}{0.32\textwidth}
    \centering
    \subcaptionbox{10-layers GCN Embedding\label{fig:k10}}{
    \includegraphics[width=0.9\textwidth]{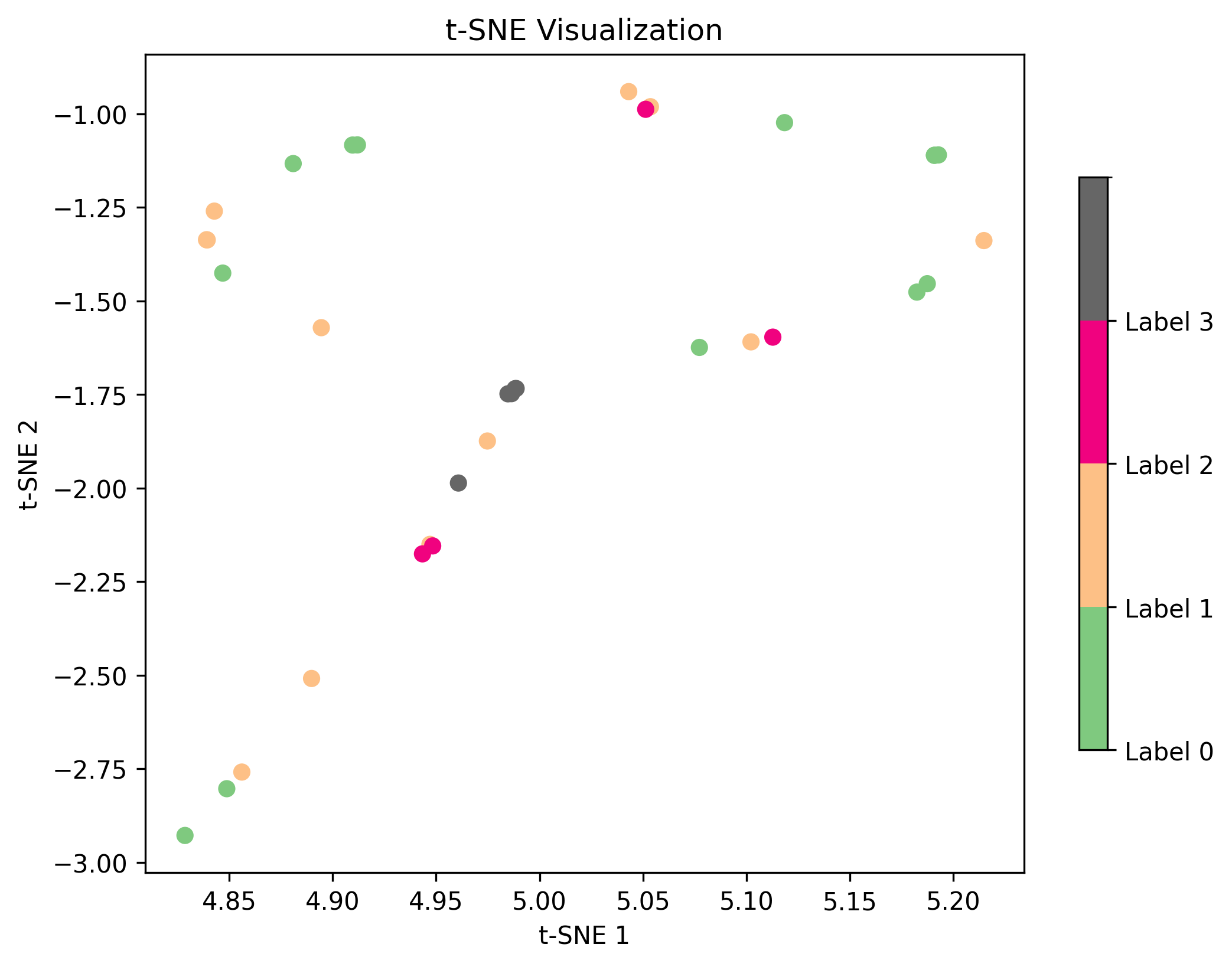}}
    \end{minipage} \hfill
    \begin{minipage}{0.32\textwidth}
    \centering
    \subcaptionbox{20-layers GCN Embedding\label{fig:k20}}{
    \includegraphics[width=1\textwidth]{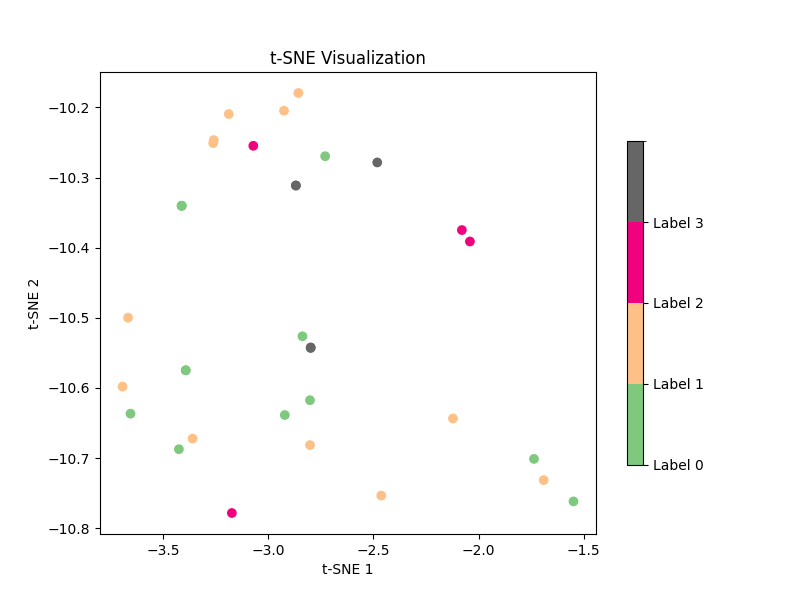}}
    \end{minipage} \hfill
    \caption{Visualization of GCN node embedding results with respect to increasing number of layers (Karate club network~\cite{karateclub}). Nodes are color-coded with same colored nodes belonging to the same class. From top to bottom, left to right, (a) denotes the Karate club network toppoloy, (b) to (f) denotes node embedding from 1-layer GCN to 20-layer GCN, respectively. As GCN layer increases from 1-layer to 3-layer, the embedding achieve better class separability (\textit{i.e.,} better results). As layer continuously, from 5-layers to 20-layers, GCN embedding loss node separability.}
    \label{fig:karateclub}
\end{figure}

\begin{figure}
    \centering
    \begin{minipage}{0.32\textwidth}
    \centering
    \subcaptionbox{1-layer GCN Embedding\label{fig:c1}}{
    \includegraphics[width=0.9\textwidth]{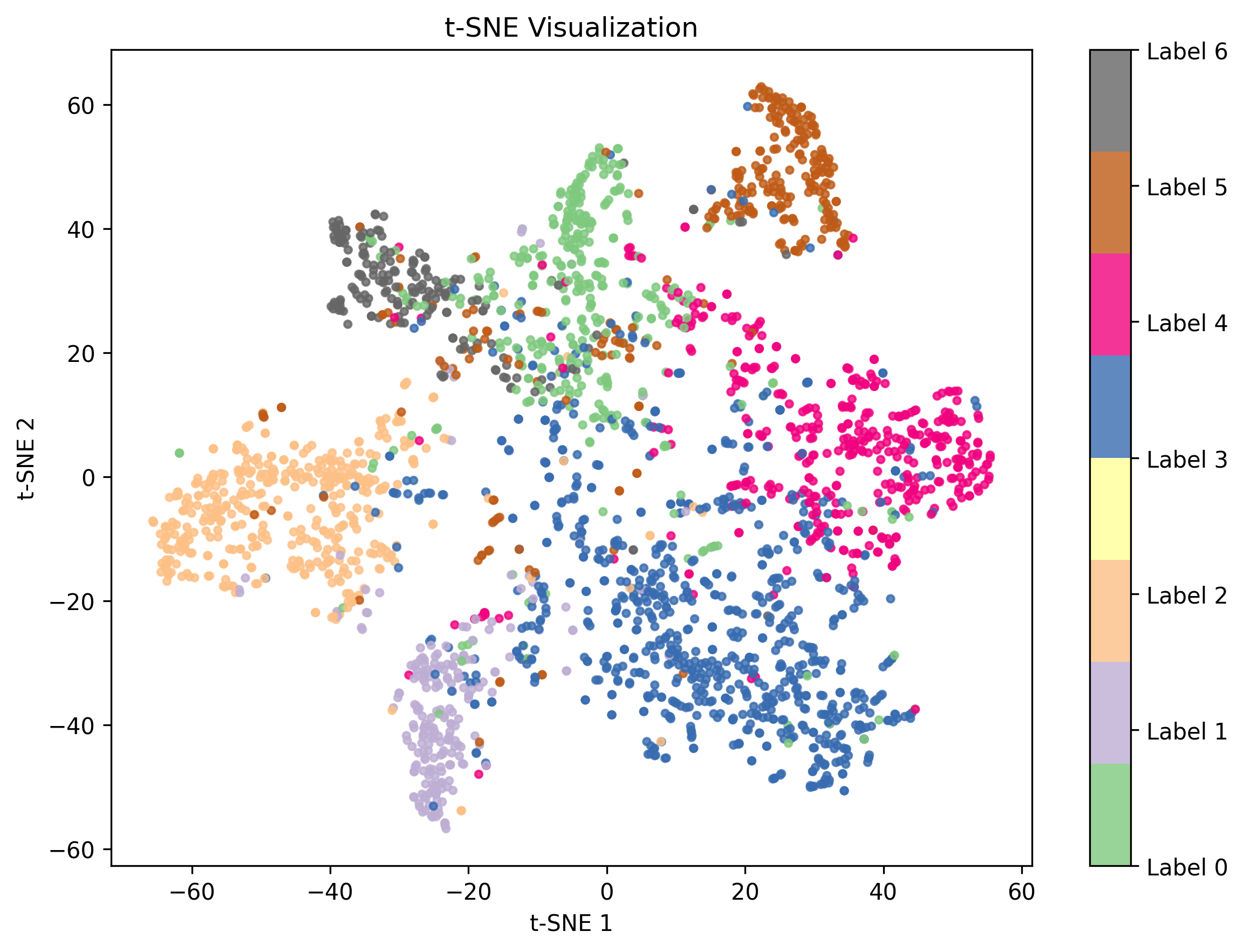}}
    \end{minipage} \hfill
    \begin{minipage}{0.32\textwidth}
    \centering
    \subcaptionbox{3-layers GCN Embedding\label{fig:c3}}{
    \includegraphics[width=0.9\textwidth]{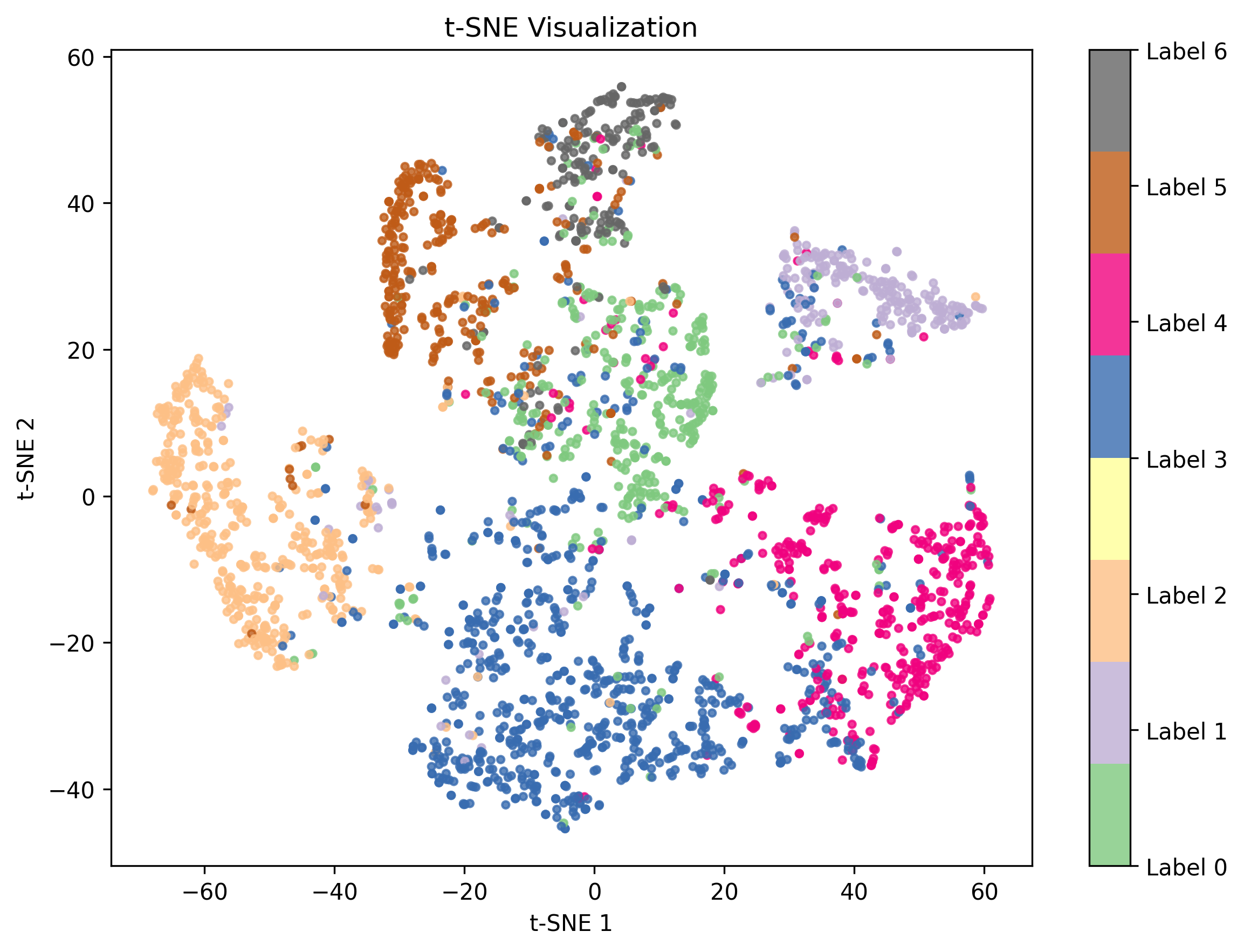}}
    \end{minipage} \hfill
    \begin{minipage}{0.32\textwidth}
    \centering
    \subcaptionbox{5-layers GCN Embedding\label{fig:c5}}{
    \includegraphics[width=0.9\textwidth]{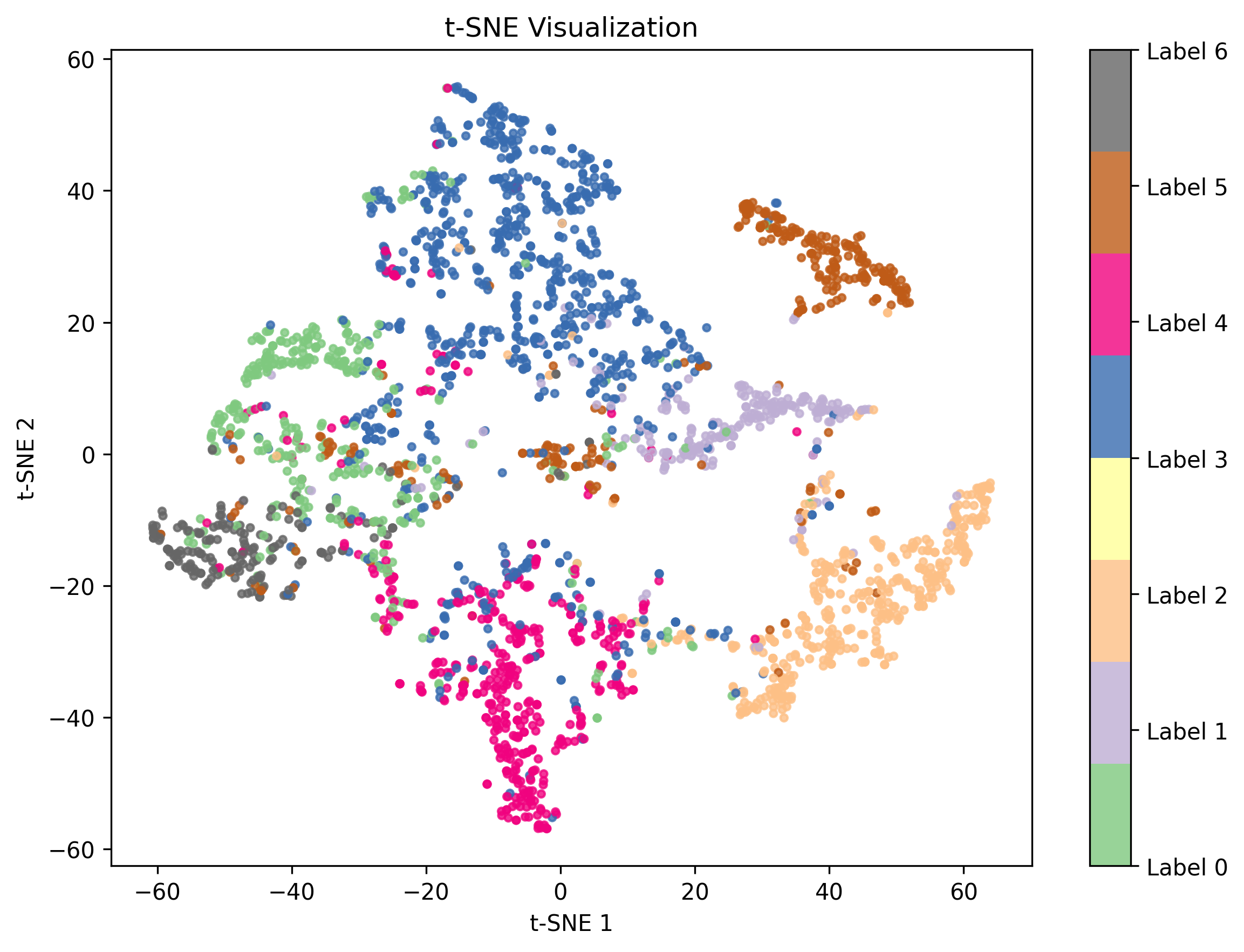}}
    \end{minipage} \hfill
    \begin{minipage}{0.32\textwidth}
    \centering
    \subcaptionbox{10-layers GCN Embedding\label{fig:c10}}{
    \includegraphics[width=0.9\textwidth]{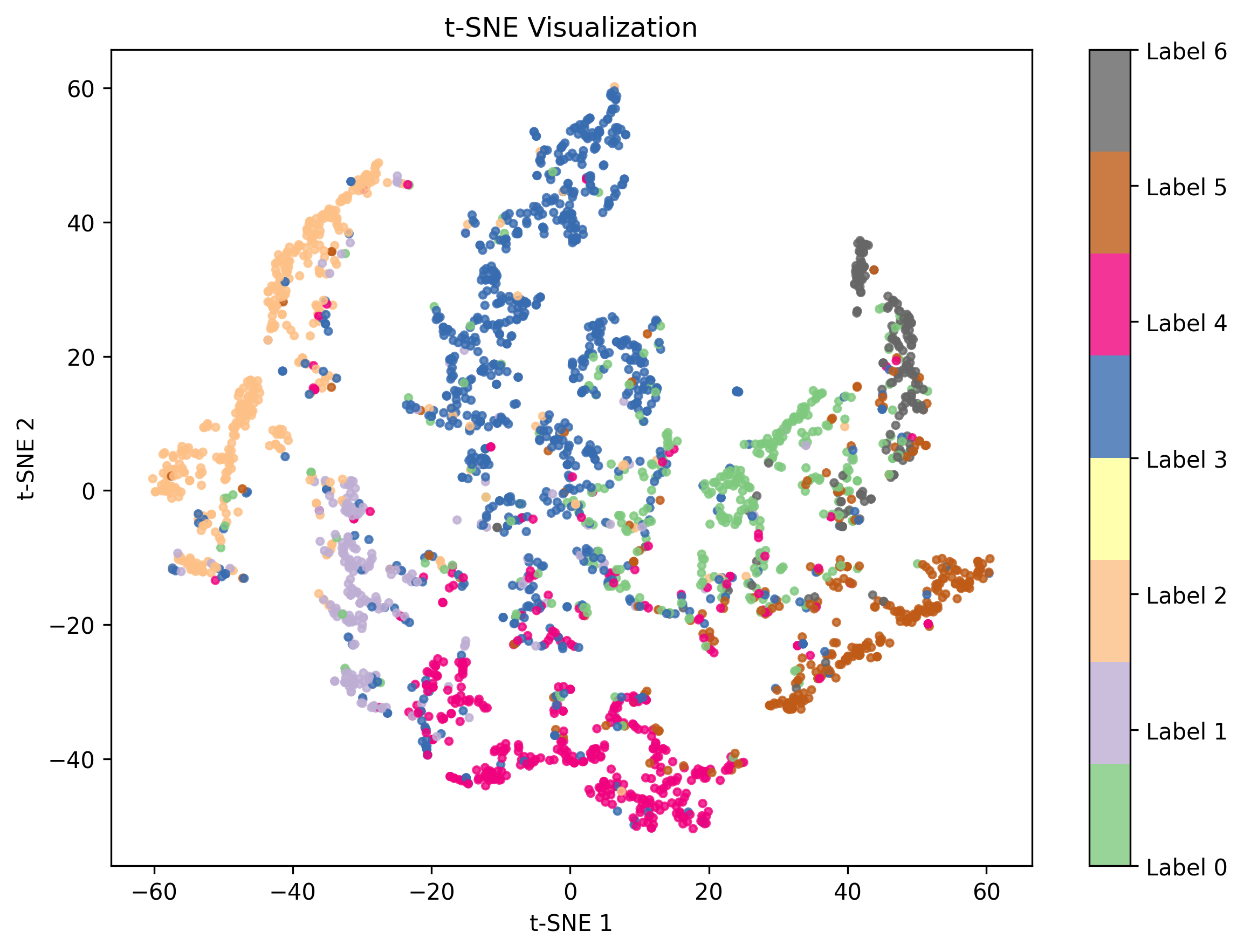}}
    \end{minipage} \hfill
    \begin{minipage}{0.32\textwidth}
    \centering
    \subcaptionbox{15-layers GCN Embedding\label{fig:c15}}{
    \includegraphics[width=0.9\textwidth]{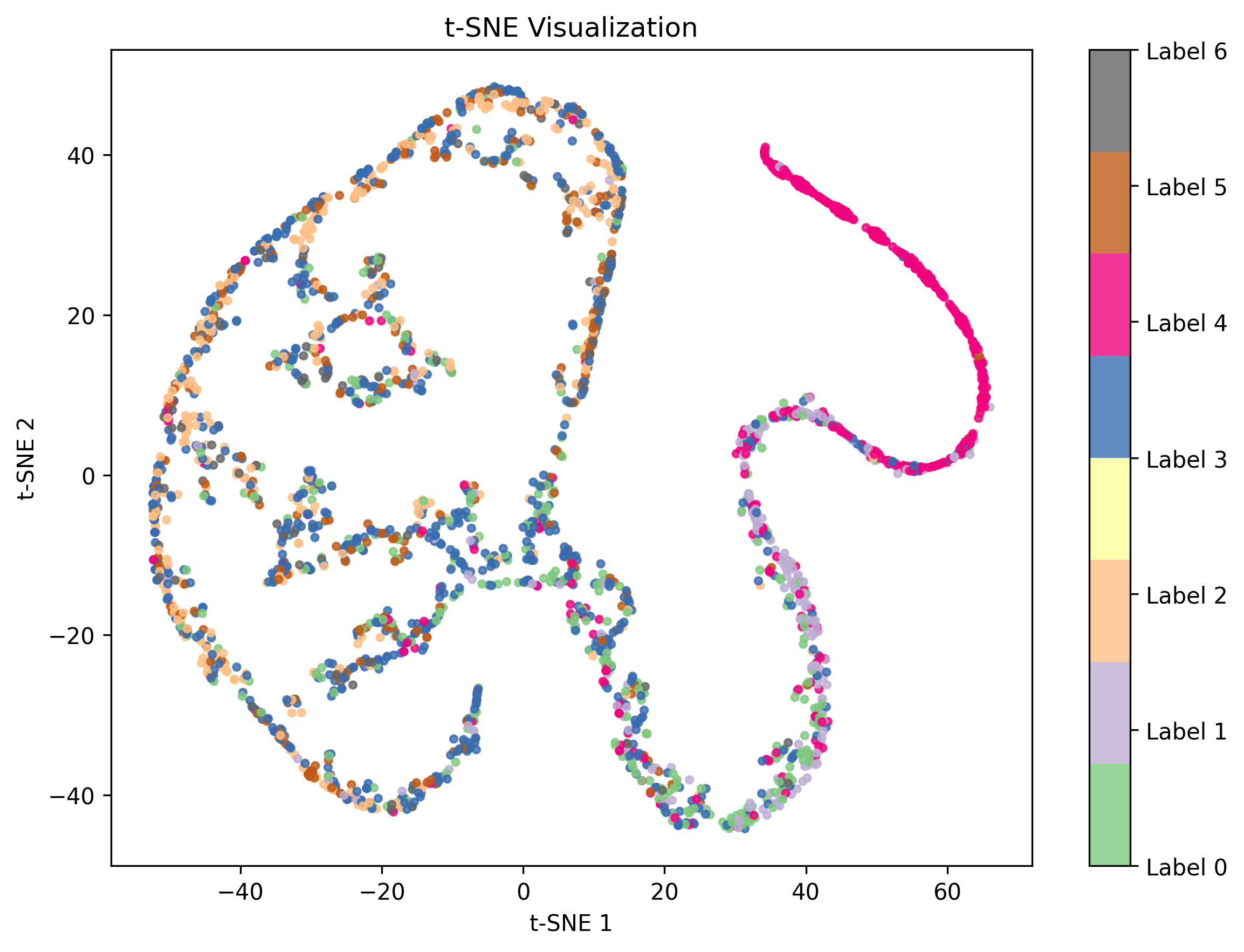}}
    \end{minipage} \hfill
    \begin{minipage}{0.32\textwidth}
    \centering
    \subcaptionbox{20-layers GCN Embedding\label{fig:c20}}{
    \includegraphics[width=0.9\textwidth]{figures/tsne-cora-layer20.png}}
    \end{minipage} \hfill
    \caption{Visualization of oversmoothing from GCN embedding learning (Cora network~\cite{Sen_Namata_Bilgic_Getoor_Galligher_Eliassi-Rad_2008}). Nodes are color-coded with same colored nodes belonging to the same class. From top to bottom, left to right, (a) to (f) denotes node embedding from 1-layer GCN to 20-layer GCN, respectively. As GCN layer increases from 1-layer to 5-layer, the embedding achieve better class separability (\textit{i.e.,} better results). As layer continuously, from 10-layers to 20-layers, GCN embedding loss node separability.}
    \label{fig:enter-label}
\end{figure}

It has been shown that simply stacking GNN layers to build a deep architecture cannot learn well due to the observed oversmoothing phenomenon~\cite{li2018deeper,NT2019RevisitingGN}. Oversmoothing can be described as a phenomenon that all node emebeddings, after deep GNN layers, become similar to each other. Several measures such as Dirichlet energy \cite{Rusch2022graphcon} and Mean Average Distances (MAD) have been proposed to quantify the extent of oversmoothing of a model \cite{mad}. In this paper, unless otherwise specified, Dirichlet energy will be used as the major measure of oversmoothing for analysis. Figure~\ref{fig:karateclub} demonstrates the GCN embedding learning results from the Karate network~\cite{karateclub}, where increasing GCN layers from 1 to 3 results in better class separability whereas increasing the layer further results in embedding features with deteriorated class separability. Figure~\ref{fig:enter-label}(b) demonstrates the oversmoothing phenomenon on the Cora network~\cite{Sen_Namata_Bilgic_Getoor_Galligher_Eliassi-Rad_2008} where all node embedding becomes similar to each other.

GNN models equipped with oversmoothing alleviation can in general accommodate more GNN layers and therefore allow nodes to have a larger receptive fields~\cite{alon2021on}. As a result, models aiming to alleviate oversmoothing also tend to gain advantage over heterophilous datasets. Such dual relation has been observed in several studies~\cite{yan2022sides,rusch2022g2gating}.

\subsection{Research Gap and Motivations}
Indeed, many works have been proposed to tackle oversmoothing in GNNs, by using different types of design principles. For example, energy-control approaches aim to increase initial energy or keep
energy from exponential decay during propagation in GNNs. Other methods focus on decoupling topology propagation and feature transformation. The vast difference in their design principles, combined with complicated graph topology and message passing mechanisms, making it difficult to understand and even compare their difference in tackling the oversmoothing. Very few survey papers exist to review methods tackling GNN oversmoothing challenges. A recent survey~\cite{survey} has compared several methods, and pointed out that some of the existing methods (such as GCNII \cite{chen20gcnii}, GraphCon \cite{Rusch2022graphcon}) cannot increase their model performance with deep layers despite the oversmoothing measure (\textit{i.e.}, Dirichlet energy) is preserved to be constant among layers, mainly because of lacking expressive power. 
Therefore, existing survey~\cite{survey} is mainly focused on reviewing method drawback from the expressive power perspective.

To date, there is no literature focusing on summarizing and comparing different alleviation approaches. Collectively, there is a missing knowledge of main themes and categorization of existing methods in the field, which may help researchers understand design principles to tackle oversmoothing. Individually, there is a lack of comparison of main stream approaches (\textit{e.g.} strength and weakness) to guide future research. 

Classical GNN message passing includes two critical operations: message aggregation and message update, where message aggregation focus on how to aggregate information passed from connected neighbors and message update focuses on how to update the node information to the next iteration or next layer. In many works, augmentation prior to training, such as feature or edge dropout, is also a part of standard pipeline, which can be simply described as a stochastic masking over original data sources, including graph topology and node features for graph learning. 

To alleviate oversmoothing for GNN learning,  most existing methods are motivated by modifying or changing the three operations, including augmentation, message aggregation, and message update, in the GNN training process with the objective of obtaining final embeddings without oversmoothing. Nevertheless, there is no systematic study of existing methods on how the three main operations affect the oversmoothing problem. To close the gap and provide guidance for future studies of oversmoothing, we propose a taxonomy with a unified view of five key steps: Augmentation, Transformation, Normalization, Propagation, and Aggregation, to summarize GNN oversmoothing alleviation approaches. The unified view therefore provides a systematic study of the three operations (augmentation, message aggregation, and message update) with more detailed analysis and categorization of existing methods. 

We notice that the three major themes only correspond to three main operations in classical GNN but also reflect the underlying different principles that are leveraged to alleviate oversmoothing. The methods we selected, despite various motivations, inspirations, and structures, all fall into the three major themes and their underlying principles.

\subsection{Contributions}
In this paper, we propose to unify existing methods in the same form and study their connections in tackling GNN oversmoothing. Our study not only provides a taxonomy and a unified view, \method\, to summarize all methods using common math formulations, but also separates them into three themes and six subgroups, by taking their unique designs into consideration. The survey outlines the differences between methods in each group, explains their rationality, and addresses their limitations. Our review has a number of math formulas, because reviewed papers are heavy in math formulations. To precisely summarize and highlight their difference, we keep representative methods' backbone formulas in the review for a better understanding. 

The remainder of the paper is structured as follows. Section 2 provides the definition of oversmoothing problem and its common measures in existing works. Section 3 describes our taxonomy for GNN oversmoothing problems from different angles. Section 4 introduces our main unified view and the six categories of existing methods, followed by a detailed discussion over each category. Section 5 concludes our works along with future guidelines for oversmoothing problems. For ease of reference, Table~\ref{tab:notations} summarizes key symbols and notations used in the paper.


\begin{table}[ht]
\begin{small}
\centering
  \caption{Summary of key symbols and notations.}
\begin{tabular}{c|p{0.8\columnwidth}}
\hline
Notations  &  Descriptions             \\ 
\hline
$G=\left (V, E, X\right)$   & An attributed graph with node set ($V$), edge set ($E$), node feature content ($X$)\\ 
$A\in\mathbb{R}^{n\times n}$   & An adjacency matrix \\
$\tilde{A}$ & An augmented or synthesized adjacency matrix\\
$\sigma(\cdot)$   & A non-linearity activation function \\
$H$, $H'$, $H''$  & Graph node embedding, the changing rate of $H$ (first derivative), and the changing rate of $H'$ (second derivative)\\
$H^l\in\mathbb{R}^{n\times f}$ & Graph node feature embedding learned at the $l^{th}$ layer\\
$H_{c}^{l}$ & Graph node feature embedding learned through convolution (or in general any local aggregation) \\
$U$ & $U$ denotes the velocity of embedding changing (assume graph node feature embedding evolves continuously with time $t$)\\
$\texttt{f}_\theta(\cdot)$ & A content-based feature encoder parameterized by $\theta$\\
$\texttt{F}_\theta(\cdot)$ & A graph convolution operator parameterized by $\theta$\\
$\texttt{NT}(\cdot)$ & Normalization techniques in GNNs\\
$\texttt{LA}(\cdot)$ & Layer-wise aggregations\\
$\texttt{Aug}(\cdot)$ & Topology augmentation function\\
\hline
\end{tabular}
\label{tab:notations}
\end{small}
\end{table}

\section{Problem Notation}
A graph with $n$ nodes is denoted by $G(V, E, X)$, where $V=\{v_1,\ldots,v_n\}$ is the vertex set with $|V|=n$, $E$ is the edge set, and $X\in\mathbb{R}^{n\times m}$ is the node content matrix recording $m$ dimensional attributes for each node. For ease of representation, we use $A\in\mathbb{R}^{n\times n}$ to denote adjacency matrix of $G$, with $A[i,j]=1$ if an edge connects $v_i$ and $v_j$, or 0 otherwise. Learning node embedding (or feature representation) is essential for graph neural networks. Meanwhile, because embedding learning is often carried out in a layer-by-layer fashion, we use $H^l\in\mathbb{R}^{n\times f}$ to denotes feature embedding learned at the $l^{th}$ layer (where each node is denoted by an $f$ dimensional latent features). $\sigma(\cdot)$ denotes a non-linearity activation function. In the following, we define operators commonly used in GNN learning and will be using these operators in the later analysis.

\vspace{0.2cm}\begin{definition}[\textbf{Feature encoders: $\texttt{f}_\theta(\cdot)$ and $\texttt{F}_\theta(\cdot)$}]
We use $\texttt{f}_\theta(\cdot)$ to denote a content based feature encoder, parameterized by learnable parameters $\theta$, converting node attributes $X$ into latent feature space. This can be achieved by using simple multi-layer perceptron (MLP) or more sophisticated learners, such as CNN or LSTM (for network having image or text as node content). Likewise, $\texttt{F}_\theta(\cdot)$ denotes graph convolution operators which leverage both node content and network topology to derive latent features. Because feature encoders typically work in a layer-wise manner, we use following notations to denote their propagation between layers.
\begin{align}
    &H^{l-1} \leftarrow  \texttt{f}_\theta(H^{l-1}) \label{eq:dense-init}; \\
    &H^{l} \leftarrow \texttt{F}_{\theta}(H^{l-1},A): H^1=X \label{eq:dense-conv}
\end{align}
\end{definition}
Where Eq.~(\ref{eq:dense-init}) represents encoding node feature only before propagation and Eq.~(\ref{eq:dense-conv}) shows the propagation process after feature encoding.

Batch normalization has been proven to be an effective component in deep neural architectures in many fields such as computer vision and natural language process. Inspired by the success of batch normalization \cite{batch} and subspace theorem \cite{subspace}, normalization techniques have been proposed to alleviate the oversmoothing in graph neural networks. 
\vspace{0.2cm}\begin{definition}[\textbf{Normalization operator: $\texttt{NT}(\cdot)$}]
We use $\texttt{NT}(\cdot)$ to denote normalization techniques in GNNs, where $\cdot$ input could be learnt embedding $H$ only or combined with topology $A$ for normalization to accommodate graph structure. An example of $\texttt{NT}(\cdot)$ is the PairNorm method \cite{pairnorm} as follows where where $H$ and $\bar{H}$ denote node embedding and its mean, $s$ is a hyperparameter, $n$ is the number of nodes, and $\|(\cdot)\|_2$ denote L2 norm.
\begin{equation}
    \texttt{NT}(H) = \frac{s\sqrt{n}(H-\bar{H})}{\|H\|_2} \label{eq:pairnorm}
\end{equation}
\end{definition}
Note that $\cdot$ input for $\texttt{NT}(\cdot)$  could be both $X$ and $A$. While normalization techniques are different, the principle behind is the same: to preserve Dirichlet energy (an important measure for oversmoothing)~\cite{Rusch2022graphcon} or to reduce the variance of the learned embeddings~\cite{nodenorm}.
 
\vspace{0.2cm}\begin{definition}[\textbf{Layer aggregator: $\texttt{LA}(\cdot)$}]
We use $\texttt{LA}(\cdot)$ to denote layer-wise aggregations that aggregate embeddings learnt from current and preceding layers. Examples of aggregation include concatenate, max pooling, and LSTM-attention operations~\cite{jknet}. 
\begin{align}
    &H^{l} \leftarrow \texttt{LA}(\cup_{i=1}^{l}H^{i}) \label{eq:dense-la}
\end{align}
\end{definition}
Notice that Eq.~(\ref{eq:dense-la}) often occurs in dense-based approach and can be treated as an ensemble trick over different layers.
\vspace{0.2cm}\begin{definition}[\textbf{Topology augmentation operator: $\texttt{Aug}(\cdot)$}]
We use $\texttt{Aug}(\cdot)$ to denote topology augmentation function using given input to generate an adjacency matrix $\tilde{A}$. For example $\texttt{Aug}({H}^l, X)$ uses node attributes $X$ and latent features at layer $l$ to generate an adjacency matrix $\tilde{A}$. 
\end{definition}

A common choice of $\texttt{Aug}(\cdot)$ could be symmetric Laplacian, Laplacian, First-order Chebyshev approximation (akin to GCN) following traditional spectral graph theory. Other choices include different random masking schemes such as random edge dropping \cite{rong2020dropedge} which is proven to be effective both empirically \cite{rong2020dropedge} and theoretically \cite{subspace}, and learnable attention matrix that has the same structure as A (examples include transformer architecture~\cite{dwivedi2021generalization} and diffusivity in GRAND \cite{chamberlain2021grand}).


\subsection{Oversmoothing Definition}
According to \cite{subspace}, oversmoothing is defined as features exponentially converging to a subspace that is invariant to the propagation matrix $A$. 
Assume $M\in\mathbb{R}^{n\times k}, k\ll n$ is a subspace invariant to $A$ (or $A$'s augmentation $\texttt{Aug}(A)$)
\textit{i.e.}, for $\forall \Omega \in M$, $A\Omega \in M$ as well. $D_M(H)$ is defined as the distance between $H$ and its closest element in $M$, \textit{i.e.},
\begin{equation}
    D_M(H) = \inf_{\Omega\in M}\|H-\Omega\|^{2}_{F}
\end{equation}
where $\|\cdot \|^{2}_{F}$ is the matrix norm. Oversmoothing indicates that $D_{M}(H^{l}) \to 0$ exponentially converges, \textit{w.r.t} the increase of layer value $l$.

Likewise, a node similarity measure $\mu$ is defined with two axioms~\cite{survey}. $\exists c \forall i \in V$ such that $V_{i} = c$ and $\mu(c) = 0$; $\mu(x+y)\leq \mu(x)+\mu(y)$, \textit{i.e.}, $\mu(\cdot)$ satisfies triangle inequality.
Then, oversmoothing is defined below with $\mu(H^{l}) \to 0$ when $l \to \infty$, where $C_{1}$ and $C_{2}$ are constants and l is the layer number.
\begin{equation}
    \mu(H^{l}) \leq C_{1}e^{-C_{2}l}
\end{equation}
The definition above is similar to the definition in \cite{subspace} with $\mu(\cdot)$ defined as $D_{M}(\cdot)$ with the measure decay rate limit to exponential decay. The second definition is often used in diffusion-based system analysis such as \cite{Rusch2022graphcon} with $\mu$ as the Dirichlet energy of the system while the first definition is often used in GNN-backbone methods such as EGNN \cite{zhou2021egnn} and DropEdge \cite{rong2020dropedge}.

\subsection{Oversmoothing Measures}
A commonly used oversmoothing measure is Dirichlet energy which can be defined as:
\begin{equation}
    \varepsilon_{\text{DE}}(H^{l}) = \frac{1}{n}\sum_{i=1}^{n}\sum_{j \in N(i)}\|H_{i}^{l}-H_{j}^{l}\|_{2}^{2} \label{eq:de-measure}
\end{equation}
where $H_{i}^{l},H_{j}^{l}$ are feature vectors reflecting nodes $L_{2}$ distance with respect to its neighbors. In practice, the specific distance metric can be treated as a learnable parameter or hyperparameter.

Two examples of using this measure include (1) the coefficient selection of EGNN based on the lower bound of $\varepsilon_{\text{DE}}(H^{l})$, and (2) G2-gating directly leveraging $\varepsilon_{\text{DE}}(H^{l})$ to compute the coefficient for each node and feature channels. We comment here that $\varepsilon_{\text{DE}}(H^{l})$ can reflect the current convergence state of the model but cannot accurately guide the model to learn correct local oversmoothing.

Another commonly used measure is Mean Average Distance (MAD) which is defined as:
\begin{equation}
    \varepsilon_{\text{MAD}}(H^{l}) = \frac{1}{n}\sum_{i=1}^{n}\sum_{j \in N(i)}\|1-\frac{(H_{i}^{l})^{T}(H_{j}^{l})}{\|H_{i}^{l}\|\|H_{j}^{l}\|}\|_{2}^{2}
\end{equation}
where we can observe that MAD simply replaced $L_{2}$ distance metric to cosine similarity compared with Eq.~(\ref{eq:de-measure}).
Note that MAD is closely related to cosine similarity and therefore only considers the direction of the two embeddings and ignores their magnitude difference. Compared with Dirichlet energy measure, caution on feature magnitude is needed when using the MAD measure.

\begin{sidewaystable}[ht]
\caption{Common dataset used for existing methods. Only datasets that are used at least twice are collected. Year reflect the time that the method is proposed. \# of Total Usage indicates the number of times the dataset is used for existing method.}
\centering
\setlength{\tabcolsep}{2pt}
\fontsize{4pt}{5pt}\selectfont
\begin{tabular}{@{}cccccccccccccccccc@{}}
\toprule
Methods/Datasets & Year & Cora~\cite{10.5555/3045390.3045396} & Citeseer~\cite{10.5555/3045390.3045396} & Pubmed~\cite{10.5555/3045390.3045396} & Texas~\cite{Pei2020Geom-GCN:} & Cornell~\cite{Pei2020Geom-GCN:} & Wisconsin~\cite{Pei2020Geom-GCN:} & CoauthorCS~\cite{shchur2019pitfallsgraphneuralnetwork} & Coauthors-Physics~\cite{shchur2019pitfallsgraphneuralnetwork} & Chameleon~\cite{10.1093/comnet/cnab014} & Amazon Photos~\cite{shchur2019pitfallsgraphneuralnetwork} & Squirrel~\cite{10.1093/comnet/cnab014} & ogbn-arxiv~\cite{10.1162/qss_a_00021}~\cite{NIPS2013_9aa42b31} & Amazon Computers~\cite{shchur2019pitfallsgraphneuralnetwork} & Film~\cite{Pei2020Geom-GCN:} & Cora-ML~\cite{McCallum2000AutomatingTC} & Reddit~\cite{10.5555/3294771.3294869} \\ \midrule
APPNP & 2018 &  & $\checkmark$ & $\checkmark$ &  &  &  &  &  &  &  &  &  &  &  & $\checkmark$ &  \\
GCNII & 2020 & $\checkmark$ & $\checkmark$ & $\checkmark$ & $\checkmark$ & $\checkmark$ & $\checkmark$ &  &  & $\checkmark$ &  &  &  &  &  &  &  \\
GEN & 2020 &  &  &  &  &  &  &  &  &  &  &  & $\checkmark$ &  &  &  &  \\
GroupNorm & 2020 & $\checkmark$ & $\checkmark$ & $\checkmark$ &  &  &  & $\checkmark$ &  &  &  &  &  &  &  &  &  \\
EGNN & 2021 & $\checkmark$ &  & $\checkmark$ &  &  &  &  & $\checkmark$ &  &  &  & $\checkmark$ &  &  &  &  \\
G2-gating & 2023 &  &  &  & $\checkmark$ & $\checkmark$ & $\checkmark$ &  &  & $\checkmark$ &  & $\checkmark$ &  &  & $\checkmark$ &  &  \\
GRN & 2024 & $\checkmark$ & $\checkmark$ & $\checkmark$ & $\checkmark$ & $\checkmark$ & $\checkmark$ &  &  & $\checkmark$ &  & $\checkmark$ &  &  & $\checkmark$ &  &  \\
JKnet & 2018 & $\checkmark$ & $\checkmark$ &  &  &  &  &  &  &  &  &  &  &  &  &  & $\checkmark$ \\
MixHop & 2019 & $\checkmark$ & $\checkmark$ & $\checkmark$ &  &  &  &  &  &  &  &  &  &  &  &  &  \\
DAGNN & 2020 & $\checkmark$ & $\checkmark$ & $\checkmark$ &  &  &  & $\checkmark$ & $\checkmark$ &  & $\checkmark$ &  &  & $\checkmark$ &  &  &  \\
DropEdge & 2020 & $\checkmark$ & $\checkmark$ & $\checkmark$ &  &  &  &  &  &  &  &  &  &  &  &  & $\checkmark$ \\
DropConnect & 2020 & $\checkmark$ & $\checkmark$ &  &  &  &  &  &  &  &  &  &  &  &  & $\checkmark$ &  \\
DropMessage & 2023 & $\checkmark$ & $\checkmark$ & $\checkmark$ &  &  &  &  &  &  &  &  & $\checkmark$ &  &  &  &  \\
PairNorm & 2020 & $\checkmark$ & $\checkmark$ & $\checkmark$ &  &  &  & $\checkmark$ &  &  &  &  &  &  &  &  &  \\
NodeNorm & 2020 & $\checkmark$ & $\checkmark$ & $\checkmark$ & $\checkmark$ & $\checkmark$ & $\checkmark$ & $\checkmark$ &  &  & $\checkmark$ &  &  &  &  &  &  \\
WeightRep & 2024 & $\checkmark$ & $\checkmark$ & $\checkmark$ &  &  &  & $\checkmark$ & $\checkmark$ & $\checkmark$ & $\checkmark$ & $\checkmark$ &  & $\checkmark$ &  &  &  \\
GRAND & 2021 & $\checkmark$ & $\checkmark$ & $\checkmark$ &  &  &  &  & $\checkmark$ &  & $\checkmark$ &  & $\checkmark$ & $\checkmark$ &  &  &  \\
Neural Sheaf Diffusion & 2022 & $\checkmark$ & $\checkmark$ & $\checkmark$ & $\checkmark$ & $\checkmark$ & $\checkmark$ &  &  & $\checkmark$ &  & $\checkmark$ &  &  & $\checkmark$ &  &  \\
GraphCon & 2022 & $\checkmark$ & $\checkmark$ & $\checkmark$ & $\checkmark$ & $\checkmark$ & $\checkmark$ &  &  &  &  &  &  &  &  &  &  \\
ACMP & 2023 & $\checkmark$ & $\checkmark$ & $\checkmark$ & $\checkmark$ & $\checkmark$ & $\checkmark$ &  & $\checkmark$ &  & $\checkmark$ &  &  & $\checkmark$ &  &  &  \\ \midrule
\# of Total Usage & \textbackslash{} & 17 & 17 & 16 & 7 & 7 & 7 & 5 & 5 & 5 & 5 & 4 & 4 & 4 & 3 & 2 & 2 \\ \midrule
\# of Nodes & \textbackslash{} & 2708 & 3327 & 19717 & 183 & 183 & 251 & 18333 & 34493 & 2277 & 7650 & 5201 & 169343 & 13752 & 7600 & 2810 & 232965 \\ \bottomrule
\end{tabular}
\label{tab:dataset}
\end{sidewaystable}

\subsection{Oversmoothing Benchmarks}
Table~\ref{tab:dataset} lists commonly used benchmark datasets by existing methods for oversmoothing validation. The datasets are listed and ordered, from left to right, based on the number of times each dataset is used in descending order. Cora, Citeseer, and Pubmed datasets are three common homophilic benchmarks used. Texas, Cornell, Wisconsin are three common heterophilic benchmarks used. Frequently used datasets are all small or medium scale ranging from $10^3$ to $10^4$ number of nodes. Few methods test shared large-scale heterophilic or homophilic benchmarks over $10^5$ number of nodes.

Oversmoothing problem occurs when GNN layers are stacked, aiming to learning global and longer hop-size node relations. For node classification tasks, homophilic graphs assume that nodes intend to share same labels with their nearby neighbors and the heterophilic graphs have more nodes sharing same labels with distant hops away. Heterophilic graphs could show more benefits and improvements for the models with deep GNN layers and reflect the model's ability to learn the global node relationship\cite{rusch2022g2gating}. Therefore, we recommend that future studies focus more on datasets that are heterophilic for testing the effectiveness of oversmoothing alleviation. Moreover, it can be observed that few large-scale heterophilic datasets are commonly tested by existing methods, resulting a lack of recognition of scalability comparison among existing methods.

\begin{table*}[tb!]
\caption{A summary of representative methods \textit{w.r.t} their categorization and properties in tackling oversmoothing.}
\resizebox{\textwidth}{!}{%
\begin{tabular}{@{}cccccccccc@{}}
\toprule
Methods & Category & Energy(Rewiring) & Energy(Nomalization)&Energy(Coefficient) &Energy(Initialization)& Decoupling&Dynamics& Residual& Dense \\ \midrule
ResGCN~\cite{li2019resgcn} & residual connection &  &  & \checkmark &  &  &  & \checkmark &  \\
APPNP~\cite{Klicpera2018appnp} & residual connection & &  & \checkmark &  & \checkmark& & \checkmark & \\
GCNII~\cite{chen20gcnii} & residual connection & &   & \checkmark &   &  & & \checkmark & \\ \hdashline
GEN~\cite{Li2020gen} & residual connection/energy control & & \checkmark &   &   &  & & \checkmark &\\
EGNN~\cite{zhou2021egnn} & residual connection/energy control & &   & \checkmark & \checkmark &  & & \checkmark &\\
GroupNorm~\cite{groupnorm} & residual connection/energy control & & \checkmark &   &   &  & & \checkmark &\\
GRN~\cite{10884161} & residual connection/energy control & & \checkmark &   &   &  & & \checkmark &\\
G2-gating~\cite{rusch2022g2gating} & residual connection/energy control/physics-inform & &   & \checkmark &   &   & \checkmark & \checkmark &\\ \hdashline
JKnet~\cite{jknet} & dense concatenation &  &   &   &   &   &  & &\checkmark\\
DAGNN~\cite{dagnn} & dense concatenation &  &   &   &   & \checkmark & & &\checkmark\\
DCGNN~\cite{guo2019dcgcn} & dense concatenation &  &   &   &   &   &  & &\checkmark\\
MixHop~\cite{mixhop} & dense concatenation &  &   &   &   &   &  & &\checkmark\\\hdashline
DropEdge~\cite{rong2020dropedge} & stochastic masking & \checkmark &   &   & \checkmark &\\
DropConnect~\cite{hasanzadeh20dropconnect} & stochastic masking & \checkmark & \checkmark & & \checkmark &\\
DropMessage~\cite{dropmessage} & residual connection/stochastic masking & \checkmark  & \checkmark  &   & \checkmark &   & & \checkmark &\\ \hdashline
PairNorm~\cite{pairnorm} & energy control &   & \checkmark &\\
NodeNorm~\cite{nodenorm} & energy control &   & \checkmark &\\
WeightRep~\cite{zhuo2024graph} & energy control &   & \checkmark & \\
\hdashline
GRAND~\cite{chamberlain2021grand} & physics-inform & \checkmark &   & & \checkmark  &   &\checkmark &\\
GraphCon~\cite{Rusch2022graphcon} & physics-inform & &   & &   &   & \checkmark &\\ 
ACMP~\cite{wang2023acmp} & physics-inform & \checkmark& \checkmark & \checkmark & & &\checkmark &\\
Neural Sheaf Diffusion~\cite{bodnar2022neural} & physics-inform & \checkmark & & & & &\checkmark &\\
GraphT\cite{dwivedi2021generalization}/GraphiT~\cite{mialon2021graphit} & graph transformer & \checkmark & & & \checkmark& & & \checkmark &\\

\bottomrule
\end{tabular}%
}

\label{tab:methods_simple}
\end{table*}

\section{A Taxonomy for GNN Oversmoothing Alleviation}
In this section, we first outline message propagation process commonly used in GNN learning (Sec 3.1), then summarize main themes to tackle oversmoothing (Sec 3.2). After that, we propose a taxonomy for GNN oversmoothing alleviation, as shown in Fig.~\ref{fig:taxonomy}. 

\subsection{GNN Message Propagation}
Deep neural architectures typically require ability to preserve long-term information passing. To achieve the goal, an inter-layer information delivery mechanism is used to regulate information passing process between layers. We briefly separate such processes into the following two subgroups.

\subsubsection{Traditional Approaches: Residual vs. Dense Connections}
Residual connection and dense connection are two common approaches to achieve inter-layer information passing. Research has shown that such simple architectures can achieve long-term information preservation, and therefore be beneficial to alleviate oversmoothing in general.

To preserve long-term information during deep layer propagation, two types of connections are commonly used in existing works, namely residual connection and dense connection:

\vspace{0.2cm}\noindent\textbf{Residual Connection} in message passing scheme can be defined in Eq.~(\ref{eq:residual}) where $\alpha$, $\beta$ can be hyperparamter constant but can also be learned.
\begin{equation}
    H^{l} \leftarrow F_{\theta}(H^{l-1},A)+\alpha H^{l-1}+\beta H^{1} \label{eq:residual}
\end{equation}
We note that Eq.~(\ref{eq:residual}) provides a choice for node embeddings update to preserve part of its original information $H^{l-1}$ and $H^{1}$ rather than entirely change to its aggregated message $F_{\theta}(H^{l-1},A)$, which typically leads to smooth update.

\vspace{0.3cm}\noindent\textbf {Dense Connection} is defined in Eq.~(\ref{eq:dense2-la}). Being dense, it implies that embeddings at the current layer $H^l$ aggregate information from all preceding layers, including $l-1, l-2,\ldots$ and so on.
\begin{align}
    H^{l} \leftarrow \texttt{LA}(\cup_{i=1}^{l}H^{i}):~~~~~ H^{l} = \texttt{F}_{\theta}(H^{l-1},A)\label{eq:dense2-la}
\end{align}
From model expressive power perspective, because dense connection can be considered as a linear version of residual connection, residual connection is more expressive in general.

\subsubsection{Complex Approaches: Dynamics and Recurrence Relation}
Recently, physics-informed approaches are proposed to first model the entire graph learning process as a continuous time process (second order partial differential equation PDE or ordinary differential equation ODE) and then use different methods to discretize continuous system, leading to a nuanced recurrence relation different from traditional GNN schemes.

A general physics-informed system (assuming a static graph) can be defined in Eq.~(\ref{eq:gds}), where $H$ is the learned node embedding and $l$ is the layer number from the model's perspective or iteration number from the solver's perspective. $H'$ and $H''$ stand for the changing rate of $H$ (first derivative) and changing rate of $H'$ (second derivative), respectively. A time $t$ variable acting as a continuous feature propagation corresponds to GNN feature propagation with layer $l$ increases.
\begin{equation}
    H'' \leftarrow \texttt{F}_\theta(H,H',H'',A,t) \label{eq:gds}
\end{equation}
Discretizaing Eq.~(\ref{eq:gds}) with different discretization schemes induces a recurrence relation similar to residual connection or dense connection but with a more complex structure. An example of discretization could be
\begin{align} \label{eq:example-discrete}
    (H^{l})' \leftarrow (H^{l-1})' +& \beta(\sigma(F_\theta(A,H^{l-1}))\nonumber\\
    -&\gamma H^{l-1}-\alpha (H^{l-1})')\\ \label{eq:example-discrete2}
    H^{l} \leftarrow H^{l-1}+&\beta(H^{l})'
\end{align}
where Eq.~(\ref{eq:example-discrete}) and Eq.~(\ref{eq:example-discrete2}) shows a common form with iterative form for both its first-derivative (embedding update rate) update equation and embedding update equation.

The principle behind the physics-informed system is that energy preserved in the physics system while the system evolving can fit into Dirichlet energy measure and a discretization method therefore keeps Dirichlet energy from exponential decay and alleviates oversmoothing accordingly.

\begin{definition}[\textbf{Message propagation operator: $\texttt{Update}(\cdot)$}]
We use $\texttt{Update}(\cdot)$ to denote an abstraction of the message propagation process in GNN learning, such as residual connection, dense connection, different recurrence relation, and implicit Euler discretization~\cite{chamberlain2021grand}, \textit{etc.} 
\label{def:update}
\end{definition}

\begin{figure}[htbp]
    \centering
    \includegraphics[width=1\textwidth, height=5cm]{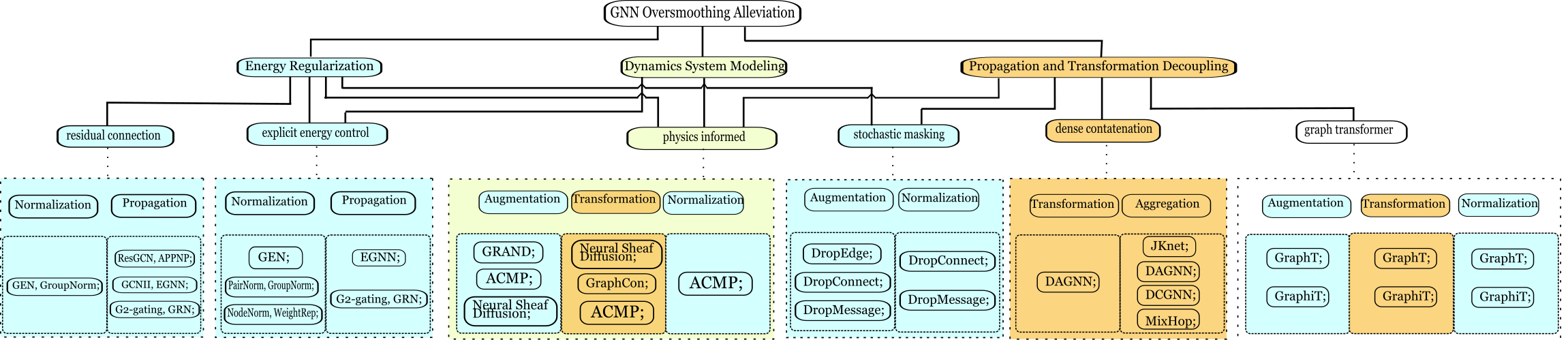}
    \caption{The proposed taxonomy of oversmooth alleviation approaches for deep graph neural networks. The taxonomy includes three themes and six categories of methods for oversmooth alleviation. Some approaches, such as physics-inform methods, are used by all three themes. Others, such as residual connection, dense concatenation, graph transformer, are only used by a unique theme.}
    \label{fig:taxonomy}
\end{figure}

\subsection{Themes to Tackle Oversmoothing}
To tackle oversmoothing, different design principles have been proposed. The themes behind these approaches are largely driven by modeling iterative GNN learning as energy regularization or as continuous system process. 
Here, the concept of energy can correspond to any oversmoothing measure or distance metric and we will refer to it as Dirichlet energy for simplicity and consistency unless otherwise specified. We summarize main themes behind existing GNN oversmoothing approaches into following three types. Table~\ref{tab:methods_simple} lists representative methods and corresponding type of approaches employed to tackle oversmoothing.

\subsubsection{Energy Regularization}
\paragraph{Initial Energy Regularization:} As defined in Sec 2.1, oversmoothing implies that the whole system energy is exponentially decayed to zero. Stochastic masking based methods provide a simple solution to alleviate the oversmoothing issue with GNN-backbone. The analysis~\cite{subspace} has shown that with a relatively less dense graph, GCN is less likely to suffer from information loss (\textit{i.e.}, oversmoothing). Therefore, DropEdge \cite{rong2020dropedge} randomly reducing the density of the graph in the beginning naturally alleviates oversmoothing. Similarly, EGNN uses orthogonal weight initialization~\cite{zhou2021egnn} to ensure each layer's initial energy is upper bounded at the starting point of training. Both methods are consistent with the analysis~\cite{subspace} that energy is related to both propagation matrix $\texttt{aug}(A)$ and learnable weight $W$. 

\paragraph{Energy Decay Regularization:} Armed with the measure of oversmoothing (Dirichlet energy), existing methods optimize the structure (\textit{e.g.}, GraphCon \cite{Rusch2022graphcon}), coefficient(\textit{e.g.}, G2-gating \cite{rusch2022g2gating}), and learned features (\textit{e.g.}, PairNorm \cite{pairnorm}) to control the energy of the generated embeddings from decaying exponentially with the layer increases. Such designs provide a theoretical assurance for embeddings to not become oversmooth. However, simply maintaining embedding energy from exponential decay does not necessarily result in a model with good performance. A recent study~\cite{survey} shows that although G2-gating, GCNII, and GraphCon have similar ability in maintaining embedding energy, as the layer increases, G2-gating enhances its model expressive power (through learned coefficients), resulting in better performance than GCNII and GraphCon.
Empirical studies and theoretical analysis are needed to deepen the understanding of a model's capability in maintaining energy \textit{vs.} expressive power. 

Methods that prevent energy from decaying exponentially can be divided into three main types: structure preserved, feature normalization preserved, and coefficient preserved. The structure preserved method mostly comes from different combinations of residual connection, dense connection, and discretization of the diffusion equation, leading to different recurrence relations of the Updating function. Feature normalization preserved method includes those normalization techniques that directly apply to features such as PairNorm, NodeNorm, etc. The coefficient preserved type is about how to select the coefficient of each residual component to preserve Dirichlet energy, including EGNN, G2-gating, etc.

\subsubsection{Dynamics System Modeling}
Instead of regulating the energy decay using normalization or other approaches, an alternative solution is to model the process as a discretized dynamic system with explicit control on system evolving and avoidance of the fixed point convergence. To this end, physics-inspired continuous systems have been leveraged as a starting point for constructing the new family of graph learning structures. The continuous systems equipped with Dirichlet energy are augmented with non-linearity and discretized with different discretization schemes, resulting in complex recurrence relations different from traditional residual and dense-based methods~\cite{Rusch2022graphcon}. Different dynamic systems provide rich properties inheriting from their continuous form analysis that traditional GNN do not have.

\subsubsection{Propagation and Transformation Decoupling}
Oversmoothing is essentially tied to the feature propagation through network topology. Another way of avoid oversmoothing is to decouple the feature learning from feature propagation. Such decoupling can be achieved through two paths: (1) positional-encoding and (2) simple stacking. Positional-encoding-based methods are mostly graph transformers where graph structure information is encoded first and then concatenated with features to feed into the transformer structure. We comment here that this type of method treats the structure as plain feature information and therefore does not involve propagation operation that causes oversmoothing. Therefore, we will not discuss this type of method in detail in this survey. The simple stacking-based method, like SGC \cite{sgc} and DAGNN\cite{dagnn}), first applies feature transformation without the adjacency matrix being involved and then applies the power of the adjacency matrix to encoded features. The final learned embedding can be summarized into a kernel or diffusion-based adjacency matrix that convolutes with encoded features. 

\subsection{GNN Oversmoothing Alleviation Taxonomy}
Based on the three themes to tackle the oversmoothing in the GNNs, we propose a taxonomy in Fig.~\ref{fig:taxonomy}, which categories all methods into three themes and six subgroups. Under the newly proposed taxonomy, some approaches, such as physics-inform methods, are used by all three themes. Whereas others, such as residual connection, dense concatenation, graph transformer, are only used by a unique theme. The taxonomy helps lay the foundation for \method, the unified view and categorization (Sec 4). In the next session, we will present unified view and categorization, and review representative methods in each category, including their key steps and relation to the \method, as well as their rationality in tackling the oversmoothing challenges. 

\section{ATNPA: Unified View and Categorization}
The three themes to tackle oversmoothing differ significantly in their principles, and such difference is even more profound in respective methods' implementation. To delve into the analysis of these seemly different approaches, a unified view \method\, with five major steps (Augmentation, Transformation, Normalization, Propagation, and Aggregation) is proposed to help review and understand how different approaches address the oversmoothing. 
\begin{align}
    &\text{Augmentation:~}&&\tilde{A} \leftarrow \texttt{Aug}(X,A) \label{eq:aug}\\
    &\text{Transformation:~}&&H^{l}_{c} \leftarrow \texttt{F}_\theta(H^{l-1},\tilde{A}) \label{eq:conv}\\
    &\text{Normalization:~}&&H^{l}_{c} \leftarrow \texttt{NT}(H^{l}_{c}) \label{eq:NT1}\\
    &\text{Propagation:~}&&H^{l} \leftarrow \texttt{NT}(\texttt{Update}(H_c^{l},H^{l-1},H^{1})) \label{eq:Update}\\
&\text{Aggregation:~}&&H^{l} \leftarrow \texttt{LA}(\cup_{i=1}^{l}H^{i}) \label{eq:dense} 
\end{align} \label{eq:unified}
The \method\ unified view, defined from Eq.~(\ref{eq:aug}) to Eq.~(\ref{eq:dense}), outlines an abstract-level framework majority GNN methods follow, with all operators being defined in previous sections. Notice that after augmentation as in Eq.~(\ref{eq:aug}), it is rare to use the original matrix $A$ unless it is in a self-supervised learning settings with multi-views of graphs utilized~\cite{10020970}. In the following, we categorize all methods into six categorizes, and review each category in details in the succeeding subsections.  
\subsection{Categorization}
Following the three major themes in Sec 3.2, we categorize existing alleviation methods based on critical changes they made compared with the vanilla GNN scheme, and link them to the \method\ unified view framework. First, we summarize its general properties and show their implicit connections. Then, we discuss specific methods within each category in detail.

Energy Regularization is one major theme we categorize for alleviating oversmoothing. Both Residual connection and Dense concatenation belong to this major theme with its unique property of alleviating energy decay thorough layer updates and they separately focus on adapting the propagation step Eq.~(\ref{eq:Update}) and aggregation step Eq.~(\ref{eq:dense}). Notice that some of the methods in this category also belongs to dynamic system modeling as the entire layer stacking process can be modeled as a dynamic system and alleviating oversmoothing can be considered as controlling the energy change in the system.
\paragraph{Residual connection:} Residual-based models explicitly add skip- or residual-connection to the \method's Propagation step at Eq.~(\ref{eq:Update}). Examples include APPNP \cite{Klicpera2018appnp}, ResGCN \cite{li2019resgcn}, GCNII, GEN \cite{Li2020gen}, EGNN, G2-gating, GRN, etc). Initial works, such as ResGCN, APPNP are inspired by residual connection in the computer vision field \cite{He2015cnnres} and only contains either skip-connection $H^{1}$ or residual-connection $H^{l-1}$. GCNII covers both skip-connection $H^{1}$ and residual connection $H^{l-1}$ to its update() function and shows better performance. However, its residual coefficient remains hyperparameter. Later, EGNN and G2-gating focus on selection of each residual coefficient under the principle of preserving Dirichlet energy among layers and generalize the residual connection trick to the gating mechanism. GRN further generalizes the constraint of Dirichlet energy to a learnable metric for better adaptation.

\paragraph{Dense Concatenation:} Dense concatenation based methods explicitly aggregate all layer embeddings into the final embeddings, which is reflected in \method's Aggregation step at Eq.~(\ref{eq:dense}). Examples include JKnet \cite{jknet}, DenseGCN~\cite{guo2019dcgcn}, MixHop \cite{mixhop}, Scattering GCN \cite{min2020scattering} and DAGNN \cite{dagnn}.

In addition to the residual-based energy change regularization, many methods focus on explicit energy control directly over the node embeddings learned and this corresponds to our Normalization step at Eq.~(\ref{eq:NT1}) and Propagation step at Eq.~(\ref{eq:Update}).

\paragraph{Energy Control:} Energy-controlled models introduce normalization techniques that control Dirichlet energy or feature variance of the learned embeddings to explicitly optimizing the measure of oversmoothing and alleviate oversmoothing. This is reflected at \method's Normalization step at Eq.~(\ref{eq:NT1}) and Propagation step at Eq.~(\ref{eq:Update}). Examples include EGNN, PairNorm, NodeNorm \cite{nodenorm}, GroupNorm \cite{groupnorm}, G2-gating.

While the two main branches above focus on regularizing the energy change thorough layer stacking, another branch of alleviating oversmoothing for energy controlling energy is to modify the initial energy (corresponding to regularize the initial condition in terms of dynamic system). This type of methods often focus on modifying our \method's Augmentation step at Eq.~(\ref{eq:aug}).
\paragraph{Stochastic Masking:} Stochastic-mask (random-mask) based models randomly mask or drop edges/nodes of the original graph, corresponding to changes in \method's Augmentation step at Eq.~(\ref{eq:aug}), and then use resulted stochastic graph for propagation. Examples include DropEdge, Drop-connect \cite{hasanzadeh20dropconnect}, DropMessage \cite{dropmessage}.

Combined from the two principles stated from with either boundary condition control and system update control (corresponding to a complete solver for PDE and ODE system), inspired from dynamic system modeling in physics, a branch of method focus on explicit construct continuous ODE or PDE related to graph diffusion and discretize it. As solving ODE or PDE requires both boundary condition (initial condition) and update equation, this type of methods falls into our \method's Augmentation step at Eq.~(\ref{eq:aug}) and Propagation step at Eq.~(\ref{eq:Update}).
\paragraph{Physics-inform Process:} Physics-informed methods first model a continuous ODE or PDE related to graph diffusion equation and then discretize the continuous equation with different discretization methods. This leads to nuanced recurrence relation and possibly a combined propagation matrix learnt from both features and topology, which corresponds to \method's Augmentation step at Eq.~(\ref{eq:aug}) and Propagation step at Eq.~(\ref{eq:Update}). Examples include GraphCon, GRAND \cite{chamberlain2021grand}, ACMP~\cite{wang2023acmp}, Neural Sheaf Diffusion \cite{bodnar2022neural}.

Finally, inspired by recently emerging transformer techniques in sequence generation problems, which can learn long-context sequence relationship, an ensemble of capturing both local relation with shallow layer GNN learning and global relation learning with transformers are studied. It is also worthnoting that transformer structure can be treated as sequence propagating on a dynamic complete graph.
\paragraph{Graph Transformer:} Transformer-based methods integrate transformer structure into GNN backbones and leverage different combination or integration to allow models to learn both long-term relation (from transformer capacity) and local relation (from GNN capacity). A recent study~\cite{min2022transformer} categorizes transformer-type models into three types: (1) Graph auxiliary Type (GA) such as GraphTrans \cite{wu2022representing} and GraphBert \cite{zhang2020graphbert}, (2) positional encoder type (PE) such as Graphormer~\cite{ying2021transformers}, and (3) improved attention matrix from graph (AT) such as GraphiT \cite{mialon2021graphit} and graphT \cite{dwivedi2021generalization}. Among the three types, PE can be considered as a decoupling of feature and topology learning, and AT types mostly resemble to GNN backbones to alleviate oversmoothing. As a result, these approaches are reflected in \method's Augmentation and Transformation steps. 

\subsection{Residual Connection Methods}
Early example of residual-based deep GNN method is APPNP \cite{Klicpera2018appnp} and GCNII \cite{chen20gcnii}, which are inspired from image field residual architecture \cite{He2015cnnres}. APPNP can be summarized as (assuming f$_{\theta}$ as a one-layer MLP):
\begin{align}
    &\tilde{A} \leftarrow \texttt{Aug}(A): ~~H^{1} \leftarrow \sigma(XW) \label{eq:APPNP-aug} \\
    &H^{l} \leftarrow (1-\alpha)\tilde{A}H^{l-1}+\alpha H^{1} \label{eq:APPNP-Update}
\end{align} \label{eq:APPNP}
Notice that Eq.~(\ref{eq:APPNP-Update}) focus solely on preserve original feature information (skip-connection) without residual connection and no non-linearity function is used for feature update.

GCNII can be summarized as:
\begin{align}
    &\tilde{A} \leftarrow \texttt{Aug}(A):~~ H^{1} \leftarrow \sigma(AXW^{1}) \label{eq:GCNII-aug}\\
    &H^{l} \leftarrow \sigma(\tilde{A}(\alpha H^{l-1}+(1-\alpha)H^{1})(\beta I+(1-\beta) W^{l})) \label{eq:GCNII-Update} 
\end{align} \label{eq:GCNII}
where its update equation Eq.~(\ref{eq:GCNII-Update}) combines both residual connection and its skip connection. The residual coefficient remains to be hyperparameter.

GEN is an extension of GCNII method and can be summarized as:
\begin{align}
    H^{l} \leftarrow \texttt{F}_\theta(H^{l-1}+H^{l}_{c}): ~~~ H^{l}_{c} \leftarrow s\cdot \|H^{l-1}\|_{2}\frac{H^{l}_{c}}{\|H^{l}_{c}\|_{2}} \label{eq:GEN-Update} 
\end{align} \label{eq:GEN}
where the regularization is inspired from batch normalization \cite{batch} and it empirically works very well. 

EGNN~\cite{zhou2021egnn} uses a slightly more complex structure that includes both skip connection and residual connection:
\begin{align}
    H^{l} \leftarrow \sigma(((1-c_{min})AH^{l-1}+\alpha H^{l-1}+\beta H^{1})W^{l})
\end{align}
where $\alpha+\beta=c_{min}$ and $c_{min}$ is a positive hyperparameter chosen to satisfy the lower bound of initial Dirichlet energy.
Its main motivation is to choose an appropriate $c_{min}$ which is an lower bound of initial Dirichlet energy to keep the Dirichlet energy in a controllable range during propagation. Therefore, EGNN is an explicit energy preserving technique compared with implicit energy control by physics-informed methods.

Similar  to EGNN, G2-gating uses simple residual GCN as a backbone with the form
\begin{equation}
    H^{l} \leftarrow (1-\sigma(\varepsilon_{DE}(F_{\theta}(H^{l-1},A)))H^{l-1}+\sigma(\varepsilon_{DE}(F_{\theta}(H^{l-1},A)))H^{l-1}F_\theta(H^{l-1},A)
\end{equation}
G2-gating's main contribution to oversmoothing lies on its controllable message dropping mechanism similar to DropMessage method. G2-gating drops message after message aggregation while DropMessage drop messages before message aggregation. Therefore G2-gating has a more controllable way to preserve Dirichlet energy. We will discuss G2-gating's message dropping mechanism, GEN and EGNN's normalization technique and coefficient selection in the energy-based model in details.

\paragraph{Discussion:} Note that all above methods fit into \method's unified view by making changes to the Propagation (Eq.~(\ref{eq:Update})) steps. In general, residual based methods have been primarily focused on learning coefficients for each component (\textit{i.e.}, coefficients for $H^l$, $H^{l-1}$, or $F_\theta(H^{l-1},A)$ \textit{etc.}). Nevertheless, there is insufficient study and theoretical analysis about the order of each residual components in terms of their position \textit{w.r.t} activation function $\sigma(\cdot)$. To date, GEN is the only work that empirically validated that order they proposed works better than GCNII.

\subsection{Dense Concatenation Methods}
Existing dense-based methods include JKnet, DAGNN, and DCGCN~\cite{guo2019dcgcn}. While Mixhop and Scattering GCN do not explicitly show oversmoothing benefits, they have a similar structure as DCGCN except that the aggregation is performed on fixed multi-hop embeddings instead of previous embeddings. JKnet provides different options over the final aggregation for layer embeddings. Here we consider the concatenate version aligned with DAGNN. Final embeddings learnt from JKnet$_{cat}$ can be summarized as: 
\begin{align}
    H^{L} \leftarrow \sum_{i=1}^{L}c_{i}H^{i}:~~~ H^{l} \leftarrow \sigma(AH^{l-1}W^{l})~\&~ H^{1} \leftarrow X \label{eq:jknetcat}
    \end{align}
where the aggregation stage is only applied to the final layer and with concatenation aggregation followed by projection. This can be described as the summation of each layer embeddings with a weight $c_{i}$ learned from the projection layer, as defined in Eq.~(\ref{eq:jknetcat}). For DAGNN, final embeddings can be summarized as (assuming one layer MLP in the beginning):
\begin{align}
    H^{L} \leftarrow \sum_{i=1}^{L}c_{i}H^{i}:~~~ H^{l} \leftarrow AH^{l-1}~\&~H^{1} \leftarrow \sigma(XW)
\end{align}
It can be observed that the two dense-based methods share similar final aggregation scheme (\textit{i.e.}, final embedding can be considered as a linear combination of layer embeddings). Yet, the embedding learnt in the intermediate layers is different. JKnet$_{cat}$ still includes learnable parameters in the middle and keep non-linearity while DAGNN removes both parts. Without non-linearity and learnable paramters, DAGNN essentially becomes a diffusion kernel based method similar to \cite{gasteiger2019gdc}. We can see that both methods fit into \method's unified view in Transformation Eq.~(\ref{eq:conv}) and Aggregation Eq.~(\ref{eq:dense}), which is the key component for dense connection based method. 

Instead of applying dense connection only to the final layer aggregation, DCGCN introduces layer aggregation at every layer in a recurrence style:
\begin{align}
    &H^{l} \leftarrow F_\theta(\cup_{i=1}^{l-1}H^{l},A)
\end{align}
which is still a variant of \method's Aggregation (Eq.~(\ref{eq:conv})) with a slight difference in the order of aggregation before convolution instead of after convolution.

\paragraph{Discussion:} We note that dense-connection can be considered as an extension of residual connection with all previous embedding being used rather than only the initial embedding or previous layer embedding. Both dense-based methods and residual-based methods can be treated as attempts of positioning residual components at different locations. However, there is a lack of theoretical analysis and empirical study comparing the two types of methods in general. It is easy to observe that ignoring non-linearity, both methods can be explained in a Markov Random Walk framework \cite{sgc}. Nevertheless, we shall point out that non-linearity is an important component for increasing model capacity and expressive power in terms of deep layers and therefore should not be discarded in analysis. 
\subsection{Stochastic-masking Methods}
Randomly dropping edges or nodes is commonly considered as an augmentation technique to avoid overfitting. It has been shown that random edge dropping is also beneficial for oversmoothing alleviation~\cite{rong2020dropedge}, where the key step is to randomly generate the adjacency matrix with a subset of edges from original edges and obtain the masked adjacency $\tilde{A}$ by
\begin{align}
    &\tilde{A} \leftarrow \texttt{Aug}(A):~~~~~ \texttt{Aug}(A) = \texttt{Bern}(p) \hadmard A \label{eq:edgedrop-aug} 
\end{align}
where $\texttt{Bern}(p)$ is a matrix filled with Bernoulli distribution elements and $p$ controls the drop rate. The motivation behind edge dropping is the subspace theorem~\cite{subspace} that indicates less connected graph leading to slow convergence of oversmoothing state. The random-mask modification fits into \method's Augmentation step Eq.~(\ref{eq:aug}).

DropConnect generalizes DropEdge to edges of each feature channels instead of edges of all features. Specifically, DropConnect create different random masked adjacency matrix $\tilde{A}$ for each features instead of one shared random masked adjacency matrix for all features. 

Similar to DropConnect, DropMessage~\cite{dropmessage} has recently been proposed to unify different masking methods including edge dropping, node dropping, and Dropout. Its key modification is:
\begin{align}
    &H^{l} \leftarrow A\tilde{H}^{l-1}W:~~ \tilde{H}^{l-1} = H^{l-1} \hadmard Bern(p)
\end{align}
where $Bern(p)$ is a feature matrix filled with Bernoulli distribution elements and $p$ controls the drop rate. Note that $\tilde{H}^{l-1}$ becomes a random variable matrix and each time $\tilde{H}^{l-1}_{ij}$ is accessed during matrix production, it will be randomized.

\paragraph{Discussion:} DropEdge prefers a shared masked adjacency matrix throughout layers instead of layer wise masking as empirically a layer-wise variant has the risk of overfitting and have additional computation cost. Additionally, Dropout method is complementary to DropEdge and applying both of them is beneficial to the model performance \cite{rong2020dropedge}. DropMessage unified them together and show
theoretical that message dropping techniques increase Shannon Entropy of propagated message compared with dropping edges, nodes or features alone, which alleviates oversmoothing. Compared with DropEdge, DropConnect which change augmentation step in \method's unified view. DropMessage can be considered as combining augmentation and normalization steps in \method's unified view.

\subsection{Energy Control Methods}
Energy-based methods share common motivation of controlling generated embeddings in each layer with constraints on either preserving Dirichlet energy or reducing feature variance. Examples of preserving Dirichlet energy include EGNN, G2-gating, PairNorm, GroupNorm, while NodeNorm reduce feature variance and WeightRep directly reparameterize the learnable parameters to allow independence of input features .

Compared with GCNII randomly searching coefficient $\alpha, \beta$ for each residual component, EGNN~\cite{zhou2021egnn} explicitly limits the coefficient searching to satisfy the lower bound of the initial Dirichlet energy and control the initialized Dirichlet energy by orthogonal weight initialization. However, the coefficient is still a scalar shared for each node and feature channels and is determined by fine tuning hyperparameters. G2-gating~\cite{rusch2022g2gating} provides a way of computing coefficients according to the graph gradient, which is essentially the Dirichlet energy and uses the gating mechanism to control features that tend to converge to stop updating and therefore avoid treating coefficient as hyperparameter. In addition, G2-gating generalizes scalar coefficient to a matrix coefficient in the shape of embedding matrix, providing fine-grained energy control. Assuming that ideal embedding is that all the nodes sharing the same labels converge to the same embedding (\textit{i.e.}, locally oversmooth) while across labels, node embeddings should be different (\textit{i.e.}, large Dirichlet energy). The gating mechanism prevents node embeddings from converging globally but also limit the local oversmoothing. Therefore, G2-gating method produces only sub-optimal solutions.

Unlike G2-gating and EGNN that have a residual-GNN backbone, PairNorm~\cite{pairnorm} normalizes the feature matrix $X$ directly according to Eq.~(\ref{eq:pairnorm}) without requiring a residual component. The theoretical analysis is based on SGC which ignores non-linearity. Similar to EGNN, PairNorm's main idea is to keep the underlying distance (such as total pairwise distance) the same, before \textit{vs.} after the layer propagation. Empirically, PairNorm alleviates oversmoothing issue but its peroformance does not improve with layer increasing. The author suggests that PairNorm may not be beneficial to standard dataset such as Cora and need a more nuanced setting (\textit{i.e.}, missing features), whereas other methods have shown performance gain in the standard dataset. 
A potential reason behind PairNorm's performance degradation, \textit{w.r.t} layer increasing, is that the normalization used in PairNorm results in less expressive power for models and therefore cannot perform well, as suggested by \cite{survey}.

GroupNorm~\cite{groupnorm} uses a simple residual-GNN backbone similar to G2-gating. Unlike G2-gating focusing on determining proper coefficients, GroupNorm normalizes the features by first assigning nodes to groups (\textit{i.e.}, clustering) and then normalizes nodes within groups to push nodes within clusters to locally oversmooth. Empirically, GroupNorm reports the results of miss features settings to validate the performance gain, which shows similar problem as in PairNorm, suggesting that normalization techniques seem to weaken the expressive power of the models with layers increasing in general.

WeightRep~\cite{zhuo2024graph} suggests that learning a proper weight can avoid oversmoothing at any layer and dynamically construct an input-dependent weight and show both empirically and theoretically that such weight can avoid oversmoothing. However, it is observed that performance degradation persists after deep layers, suggesting other factors that might cause the performance degradation.
\paragraph{Discussion:} We note that both normalization and coefficient computation approaches fit into \method's unified view in Normalization Eq.~(\ref{eq:NT1}) and Propagation Eq.~(\ref{eq:Update}). Direct normalization on features such as mean substraction and variance shifting empirically reduce model capacity and expressive power while coefficients learning seem to be a more promising direction to not only keep Dirichlet energy but also preserve model expressive power.

\subsection{Physics-inform Process}
Physics-informed methods consider GNN learning as a continuous system and derive solutions by formulating the system's evolving as a model propagation process. In this context, the time component $t$ in continuous system corresponds to GNN based model's layer concept. Different discretization methods provide a complex family of methods indicated by a continuous system and most GNN based backbone can be considered as an explicit Euler discretization (only considering the recurrence relation or the Update function in GNN framework) \cite{chamberlain2021grand}. Examples of continuous systems include GRAND, GraphCon, ACMP, Neural Sheaf Diffusion, and G2-gating (It was first reviewed as GNN backbones, but is also related to the continuous system).

GRAND~\cite{chamberlain2021grand} leverages graph diffusion PDE equations as the continuous system and performs both explicit and implicit Euler discretization. The diffusivity is modeled with an attention structure Eq.~(\ref{eq:GRAND-diffusion}) related to node features and edges Eq.~(\ref{eq:GRAND-rewire}):
\begin{align}
    &\texttt{Aug}(X) \leftarrow \sigma(\frac{(KX)^{T}(QX)}{d_k}) \label{eq:GRAND-diffusion}\\
    &\texttt{Aug}(X,A) \leftarrow (\texttt{Aug}(X)>\epsilon) \hadmard A \label{eq:GRAND-rewire}
\end{align}
where $\sigma(\cdot)$ is a non-linearity activation softmax function. $K$ and $Q$ are learnable parameters, and $d_k$ is the hidden dimension for $K$ which is used as normalization.  $\epsilon$ is a threshold value to sparsify attention matrix $\texttt{Aug}(X)$ and $\hadmard$ denotes element-wise multiplication. Eq.~(\ref{eq:GRAND-diffusion}) is the diffusion variant and Eq.~(\ref{eq:GRAND-rewire}) is rewired variants for GRAND. With both discretization, the key component fits into \method's unified view in Augmentation Eq.~(\ref{eq:aug}).

Similar to GRAND, ACMP~\cite{wang2023acmp} modifies the graph diffusion equation to a particle interaction system. It generalizes GRAND's \texttt{Aug}() in Eq.~(\ref{eq:GRAND-rewire}) by adding a negative constant to the attention weight so that the attention could be negative. This allows the nodes to not only attract but also repulse each other through learning. Additionally, to control the upper bound Dirichlet energy, a well-shaped function (called double-well potential) is added as a regularization (equivalent to feature normalization) to avoid infinite Dirichlet energy growth. It fits into
\method's Augmentation and Normalization steps, despite a very different origin (particle system interpretation). 

GraphCon~\cite{Rusch2022graphcon} leverages a graph dynamic system of non-linear ODEs:
\begin{align}
    &U' \leftarrow F_\theta(A,H,t)-\gamma H- \alpha U \\
    &H' \leftarrow U 
\end{align}
where $H'$ is the first order derivative with respect to time $t$ (a default setting at physics) and $U'$ is equivalent to $H''$. After discretization, $t$ is essentially equivalent to layer $l$ in GNN backbones. Following IMEX (implicit-explicit) time discretization \cite{imex}, GraphCon obtains a new recurrence:
\begin{align}
    U^{n} \leftarrow U^{n-1}&+\Delta(t)(\sigma(F_\theta(A,H^{n-1},t^{n-1}))\nonumber\\
    &-\gamma H^{n-1} -\alpha U^{n-1})\\
    H^{n} \leftarrow H^{n-1}&+\Delta(t)U^{n}
\end{align}
where $\Delta(t)$ is the discretization step.

\paragraph{Discussion:} 
Diffusion systems above share a common point of establishing a connection between the feature changing rate $H'$ and the graph gradient $\sum_{j\in Neighbor(i)}|h_{i}-h_{j}|$ which is Dirichlet energy for one node. A discretization of the system then provides a unique complex recurrence relation that preserves established connections. 

The niche of diffusion-based methods stem from the design that the system preserves Dirichlet energy (mitigates oversmoothing) through the complex residual recurrence structure, avoiding fixed point convergence at exponential rate and small perturbation deviates the fixed point away in the GraphCon case. This makes diffusion-based method unique, compared with other works that preserve energy through explicit feature value control or coefficient control.


Neural Sheaf diffusion is an approach using cellular sheaf theory to model evolving of the features at each layer and the geometry of the graph~\cite{bodnar2022neural}.
The augmentation to Sheaf Diffusion, similar to GCN augmentation, 
constructs a continuous differential equation as:
\begin{equation}
    (H^{t})' \leftarrow -\sigma(\texttt{Aug}_{\theta}(A,H^{t})W^{t}_{1}H^{t}W^{t}_{2}) \label{eq:sheaf}
\end{equation}
where $H^{t}$ unlike common feature matrix with dimension $n \times d$ with $d$ as hidden dimension, each node feature is vertically stacked and $H^{t}$ is of dimension $nd \times 1$. $\texttt{Aug}_{\theta}(A,H^{t})$ also produces an $nd \times nd$ matrix with $n\times n$ subblocks of dimension $d \times d$. 
The discrete version of Eq.~(\ref{eq:sheaf}) becomes
\begin{equation}
    (H^{t}) \leftarrow H^{t-1}-\sigma(\texttt{Aug}_{\theta}(A,H^{t-1})W^{t-1}_{1}H^{t-1}W^{t-1}_{2}) \label{eq:sheafd}
\end{equation}
\paragraph{Discussion:} We comment here that the extra $nd$ dimensions provide each feature channel with a possibly different propagation channel compared with the original settings where the propagation channel binds to the node level. The idea behind is similar to G2-gating where they also have a multi-rate coefficient matrix to control the update fine-grained to each feature of each node instead of each node. Another point about Sheaf Diffusion is that they use shared weight among each block, \textit{i.e.} $W^{t-1}_{1}$ can be decomposed as the Kronecker product of the Identity matrix and a learnable $W^{'}_{1}$ with dimension $d \times d$ and therefore reduce the exponential number of parameter increase, which in term indicates an assumption that one feature correlation is shared among graph topology.

\subsection{Transformer-based Methods}
Transformer has shown superior performance in long-term relation learning \cite{rong2020selfsupervised}. GNN has been proven to be effective on local relation learning and performance deteriorates when both global and local relation exists (\textit{i.e.}, graphs with middle homophily scores \cite{luan2023when}. Combining transformer and GNN architecture has been used to capture both long and short-term relations and improve model performance on heterophilous graphs. With the connection between heterophily and oversmoothing \cite{yan2022sides}, we consider transformer-based methods candidates for alleviating oversmoothing. 

There are mainly three types of approaches according to the position of the two components, PE (positional encoding), GA (graph auxiliary), and AT (attention matrix from graph). Admittedly, many graph transformers simultaneously use several techniques. To understand the role each part plays in the learning process, we will discuss each component individually and fit them into the proposed framework. PE-type can be roughly considered as projecting certain graph properties to feature space and then aggregating the projected graph features with node features. The aggregated feature is then fed into the transformer block. A general form of PE for one transformer block is therefore:
\begin{align}
     X \leftarrow \texttt{Transformer}(X,\tilde{A}) :~~~ \tilde{A} \leftarrow \texttt{Aug}(A)
\end{align}
\paragraph{Discussion about PA:} Unlike normal graph convolution, $\texttt{Transformer}()$ can be considered as a complex feature transformation, where $X$ and $\tilde{A}$ are not convoluted but are processed in a transformer style. Because it avoids convolution directly, it can considered as a decoupled feature and topology learning. Theoretical analysis of model expressive power between transformer and graph convolution is lacking in existing research and potential analysis is necessary to justify the learnability of such structure.

\paragraph{Discussion about GA:} As GA type stacks transformer block with graph convolution block, making it hard to analyze and the concept of oversmoothing becomes vague in this case. Briefly speaking, we can treat the transformer block as a complex feature Transformation or Normalization steps of \method. Then GA-type transformer can be treated as a common GNN framework with complex normalization applied in the middle. Since the transformer does not ensure the energy of learned node embeddings, the GA-based component will not necessarily help alleviate oversmoothing.

AT-type graph transformers, such as GraphT and GraphiT, have unique $\texttt{Aug}(\cdot)$ components, defined in Eq.~(\ref{eq:GraphT}) for GraphT and Eq.~(\ref{eq:graphiT}) for GraphiT, which share a striking similarity to attention-based diffusivity structures, such as GRAND and ACMP.
\begin{align}
    \texttt{Aug}(X,A) \leftarrow \sigma(\frac{(XQ)(XK)^{T}}{d_{k}}\hadmard A) \label{eq:GraphT} \\
    \texttt{Aug}(X,A) \leftarrow \sigma(\frac{(XQ)(XK)^{T}}{d_{k}}\hadmard \kappa(A)) \label{eq:graphiT} 
\end{align}
where $\sigma$ is the softmax nonlinear activation function, $d_{k}$ is the hidden dimension of $X$, and $\kappa (A)\in\mathbb{R}^{n\times n}$ denotes a transformation of $A$, such as graph Laplacian. 

\paragraph{Discussion about AT:} The main difference between GraphT Eq.~(\ref{eq:GraphT}) and GRAND Eq.~(\ref{eq:GRAND-rewire}) is the location of the non-linearity activation function $\sigma(\cdot)$. The similarity between GRAND and transformer-based methods comes from the diffusivity modeled as an attention structure in GRAND and the diffusivity can fit into the $\texttt{Aug}(\cdot)$ Augmentation step in \method. AT-type methods can be treated as a graph rewiring approach. Since the rewired graph will be more sparse compared with the original graph and the rewired process can be done in each layer, we can consider them as an extension of a controllable masking mechanism aiming to increase the energy at each layer.

\section{Experimental Results Reported From Existing Methods Under Same Settings}

\begin{table*}[htbp]
\centering
\caption{Collected average results from each method following the same setting for six most frequently used datasets with three homophilic and three heterophilic datasets. For the Planetoid dataset (Cora, Citeseer, Pubmed), NA indicates missing reports. Some of the results are reported by other method's paper as its original paper has a different split setting. Bold font indicates best results and italic font indicate second best results.}
\begin{tabular}{@{}cccccccc@{}}
\toprule
Method/Dataset & Year & Cora & Citeseer & Pubmed & Texas & Cornell & Wisconsin \\ \midrule
APPNP & 2018 & 83.3 & 71.8 & 80.1 & 65.41 & 73.51 & 69.02 \\
\textbf{GCNII} & 2020 & \textit{85.5} & \textbf{73.4} & 80.2 & 77.84 & 76.49 & 81.57 \\
GroupNorm & 2020 & 82 & 69.5 & 79.5 & NA & NA & NA \\
\textbf{EGNN} & 2021 & \textbf{85.7} & NA & 80.1 & NA & NA & NA \\
\textbf{G2-gating} & 2023 & NA & NA & NA & \textit{87.57} & \textbf{87.3} & \textit{\textit{87.65}} \\
\textbf{GRN} & 2024 & NA & NA & NA & \textbf{89.73} & \textit{86.22} & \textbf{88.4} \\
JKnet & 2018 & 83.3 & 72.6 & 79.2 & 57.3 & 61.08 & 50.59 \\
MixHop & 2019 & 81.9 & 71.4 & \textbf{80.8} & 77.83 & 73.51 & 75.88 \\
DAGNN & 2020 & 84.4 & 73.3 & 80.5 & NA & NA & NA \\
DropEdge & 2020 & 81.69 & 71.43 & 79.06 & NA & NA & NA \\
DropConnect & 2020 & 82.2 & 71.72 & NA & NA & NA & NA \\
DropMessage & 2023 & 83.33 & 71.83 & 79.2 & NA & NA & NA \\
PairNorm & 2020 & 82.1 & 69.6 & 78 & 60.27 & 58.92 & 48.43 \\
NodeNorm & 2020 & 83 & 72.9 & \textit{80.7} & 78.92 & 80.54 & 83.14 \\
WeightRep & 2024 & 82.21 & 69.04 & 77.68 & NA & NA & NA \\
GRAND & 2021 & 84.7 & 73.3 & 80.4 & NA & NA & NA \\ \bottomrule
\end{tabular}

\label{tab:collected-empirical-results}
\end{table*}

To verify the effectiveness of existing methods, we report each method's best report results under similar/same settings on public datasets including three most frequently used homophilic and three heterophilic datasets. 
The results are shown in Table~\ref{tab:collected-empirical-results}. For each dataset, the best performance is bold-faced, and the second best performance is in italic format. In the table, we also highlight four methods with bold fonts that achieve significant performance boost over either homophilic datasets and heterophilic datasets, including GCNII, EGNN, G2-gating, and GRN. It is worth noting that all methods are residual based with three methods focus on limiting residual coefficients based on dirichlet energy constraint or learnable metric constraint. This result suggests a promising direction following the principle of energy change regularization over propagation and residual-based GNN structure. Additionally, we observe that GRAND also shows good performance gain with its physics-informed structure, suggesting the future of dynamic system based approaches.

\section{Conclusion}
In this paper, we reviewed and analyzed GNN oversmoothing alleviation methods. We argued that despite of dramatic differences in their design principles and math formulations, existing approaches share three common themes in their motivations to tackle oversmoothing, and the commonality allows us to summarize them into six categories. To allow in-depth understanding and analysis of all methods, we proposed \method, which uses five steps to distill properties and architectures of existing methods and shows that existing oversmoothing alleviation methods are variants by introducing changes to one or multiple steps of the \method. The unified view allows a clear understanding on how oversmoothing is alleviated for individual methods, strength and weakness of each type of methods, and possible future study directions. We drew discussion and remarks on representative methods, and observed that despite many methods focusing on constraining energy of the learned embeddings, diffusion-based methods use a physics-inspired structure to keep energy, while residual-based methods use a simple structure but focus on tuning coefficient or directly applying normalization to features to preserve energy. In addition, the modeling of network propagation has evolved from a static topology to dynamically learn topology, or to seek respective adjacency matrix for each feature instead of one shared topology for all features.

The paper summarizes three main themes according to the underlying principles. With the theme dynamic system modeling combine two principles behind energy regularization and propagation and transformation decoupling. It is possible that our paper may overlook some works that are not explicitly focused on oversmoothing. Nevertheless, we are confident that reviewed methods cover major oversmoothing alleviation branches and the underling principles leveraged are coherent to the problem and are discussed to the best of our knowledge.

From a high abstract level, we highlight that oversmoothing alleviation is equivalent to dynamic system control with constraints. It is therefore suggested in the future studies to focus on two main principles for system controlling: energy initialization or boundary condition control and energy change control thorough propagation. Moreover, our \method~ view provides an excellent component-wise separation for future researchers to modify according to the highlighted two principles. Practically, we notice that existing methods lack a unified comparisons for large-scale benchmark on both heterophilic and homophilic datasets. Scalability of existing methods could therefore become an important future study for real-world applications.

\backmatter


\bmhead{Acknowledgements}
This work has been supported in part by the US National Science Foundation (NSF) under
Grant Nos. IIS-2236579, IIS-2302786, and IOS-2430224.

\section*{Declarations}

\begin{itemize}
\item Funding: This work has been supported in part by the US National Science Foundation (NSF) under
Grant Nos. IIS-2236579, IIS-2302786, and IOS-2430224.
\item Conflict of interest/Competing interests: The authors have no conflicts of interest to declare that are relevant to the content of this article.
\item Ethics approval and consent to participate: N/A
\item Consent for publication:
\item Data availability: 
\item Materials availability:
\item Code availability: 
\item Author contribution: Y. Jin and X. Zhu made equal contribution to the article.
\end{itemize}

\noindent

\bigskip
\bibliography{ijcai24}

\end{document}